\documentclass[journal]{IEEEtran}

\usepackage{times}
\usepackage{epsfig}
\usepackage{graphicx}
\usepackage{amsmath}
\usepackage{amssymb}
\usepackage{booktabs, multicol, multirow}
\usepackage{array}
\usepackage{color}
\usepackage{subfigure}
\usepackage{caption}
\usepackage{cite}
\usepackage{balance}

\newcommand{\etal}{\textit{et al}.}
\newcommand{\comment}[1]{\textcolor{black}{#1}}
\newcommand{\commentt}[1]{\textcolor{black}{#1}}
\newcommand{\commentAQ}[1]{\textcolor{black}{#1}}

\begin{document}
%
\title{Deep Hashing for Secure \\Multimodal Biometrics}
%
%
%

\author{Veeru~Talreja,~\IEEEmembership{Student Member,~IEEE,}
        Matthew~Valenti,~\IEEEmembership{Fellow,~IEEE,}
        and~Nasser~Nasrabadi,~\IEEEmembership{Fellow,~IEEE}}

%
%

\markboth{IEEE TRANSACTIONS ON INFORMATION FORENSICS AND SECURITY}%
{Shell \MakeLowercase{\textit{et al.}}: Bare Demo of IEEEtran.cls for IEEE Journals}
%



\maketitle

\begin{abstract}
 When compared to unimodal systems, multimodal biometric systems have several advantages, including lower error rate, higher accuracy, and larger population coverage. However, multimodal systems have an increased demand for integrity and privacy because they must store multiple biometric traits associated with each user. In this paper, we present a deep learning framework for feature-level fusion that generates a secure multimodal template from each user's \commentAQ{face and iris biometrics}. We integrate a deep hashing (binarization) technique into the fusion architecture to generate a robust binary multimodal shared latent representation. Further, we employ a hybrid secure architecture by combining cancelable biometrics with secure sketch techniques and integrate it with a deep hashing framework, which makes it computationally prohibitive to forge a combination of multiple biometrics that passes the authentication. The efficacy of the proposed approach is shown using a multimodal database \commentAQ{of face and iris} and it is observed that the matching performance is improved due to the fusion of multiple biometrics. Furthermore, the proposed approach also provides cancelability and unlinkability of the templates along with improved privacy of the biometric data. Additionally, we also test the proposed hashing function for an image retrieval application using a benchmark dataset. \commentAQ{The main goal of this paper is to develop a method for integrating multimodal fusion, deep hashing, and biometric security, with an emphasis on structural data from modalities like face and iris. The proposed approach is in no way a general biometrics security framework that can be applied to all biometrics modalities, as further research is needed to extend the proposed framework to other unconstrained biometric modalities.}
 
 
 
 
\end{abstract}


%
\IEEEpeerreviewmaketitle

\section{Introduction}
%
%
%
%
\IEEEPARstart{B}{iometrics} are difficult to forge, and unlike in traditional password-based access control systems, they do not have to be remembered. As much as these characteristics provide an advantage, they also create challenges related to protecting biometrics in the event of identity theft or a database compromise as each biometric characteristic is distinct and cannot be replaced by a newly generated arbitrary biometric. There are serious concerns about the security and privacy of an individual because of the proliferation of biometric usage. These concerns cannot be alleviated by using conventional cryptographic hashing as in case of alpha-numeric passwords because the cryptographic hashes are extremely sensitive to noise and are not suitable for the protection of biometrics due to inherent variability and noise in biometric measurements. 

The leakage of biometric information to an adversary constitutes a serious threat to security and privacy because if an adversary gains access to a biometric database, he can potentially obtain the stored user information. The attacker can use this information to gain unauthorized access to the system by reverse engineering the system and creating a physical spoof. Furthermore, an attacker can abuse the biometric information for unintended purposes and violate user privacy \cite{nagar_multibiometriccryptosystems_2012}. 

Multimodal biometric systems use a combination of different biometric traits such as face and iris, or face and fingerprint. Multimodal systems are generally more resistant to spoofing attacks \cite{ross_multibiometric_2004}. Moreover, multimodal systems can be made to be more universal than unimodal systems, since the use of multiple modalities can compensate for missing modalities in a small portion of the population. Multimodal systems also have an advantage of lower error rates and higher accuracy when compared to unimodal systems \cite{nagar_multibiometriccryptosystems_2012}. Consequently, multimodal systems have been deployed in many large scale biometric applications including the FBI's Next Genration Identification (NGI), the Department of Homeland Security's US-VISIT, and the Government of India's UID. However, Multimodal systems have an increased demand for integrity and privacy because the system stores multiple biometric traits of each user. Hence, multimodal template protection is the main focus of this paper. 

The fundamental challenge in designing a biometric template protection scheme is to manage the intra-user variability that occurs due to signal variations in the multiple acquisitions of the same biometric trait. With respect to biometric template protection, four main architectures are widely used: \emph{fuzzy commitment, secure sketch, secure multiparty computation, and cancelable biometrics} \cite{rane_secure_biom_2013}. Fuzzy commitment and secure sketch are biometric cryptosystem methods and are usually implemented with error correcting codes and provide information-theoretic guarantees of security and privacy (e.g., \cite{Sutcu_2007_Protecting,Juels_1999_Fuzzy,Juels_2002_Vault,Nandakumar_2007_Fingerprint_FV,Nagar_2008_Securing}). Secure multiparty computation architectures are distance based and use cryptographic tools. Cancelable biometrics use revocable and non-invertible user-specific transformations for distorting the enrollment biometric (e.g.,\cite{ratha_canc_biom_2007,kong_analysis_2006,zuo_cancelable_2008,teoh_cancellable_2008}),  with the matching typically performed in the transformed domain. 


For a template to be secure, it must satisfy the important properties of \emph{noninvertibility} and \emph{revocability}. Noninvertibility implies that given a template, it must be computationally difficult to recover the original biometric data from the template. Revocability implies that if a template gets compromised, it should be possible to revoke the compromised template and generate a new template using a different transformation. Moreover, it should be difficult to identify that the new template and the old compromised template are generated from the same underlying biometric data. 


One important issue for multimodal systems is that the multiple biometric traits generally do not have the same feature-level representation. Furthermore, it is difficult to characterize multiple biometric traits using compatible feature-level representations, as required by a template protection scheme \cite{nagar_multibiometriccryptosystems_2012}. To counter this issue there have been many fusion techniques for combining multiple biometrics \cite {sutcu_secure_2007,nandakumar_multibiometric_2008,nagar_multibiometriccryptosystems_2012}. One possible approach is to apply a separate template protection scheme for each trait followed by decision-level fusion. However, such an approach may not be highly secure, since it is limited by the security of the individual traits. This issue motivated our proposed approach of using multimodal biometric security to perform a joint feature-level fusion and classification.

Another important issue is that biometric cryptosystem schemes are usually implemented using error control codes. In order to apply error control codes, the biometric feature vectors must be quantized, for instance by binarizing. One method of binarizing the feature vectors is thresholding the feature vectors, for example, by thresholding against the population mean or thresholding against zero. However, thresholding causes a quantization loss and does not preserve the semantic properties of the data structure in Hamming space. In order to avoid thresholding and minimize the quantization loss, we have used the idea of \emph{hashing} \cite{gionis_1999_similarity,iterative_gong_2013}, which is used in the image and data retrieval literature to achieve fast search by binarizing the real-valued image features. The basic idea of hashing is to map each visual object into a compact binary feature vector that approximately preserves the data structure in the original space. Owing to its storage and retrieval efficiency, hashing has been used for large scale visual search and image retrieval.


Recent progress in image classification, object detection, face recognition, speech recognition and many other computer vision tasks demonstrates the impressive learning ability of \emph{convolutional neural networks} (CNN). The robustness of features generated by the CNN has led to a surge in the application of deep learning for generating binary codes from raw image data. Deep hashing \cite{supervised_xia_2014,simultaneous_lai_2015,learning_lin_2016,deep_hashing_liu_2016} is the technique of integrating hashing and deep learning to generate compact binary vectors from raw image data. There is a rich literature related to the application of optimized deep learning for converting the raw image data to binary hash codes. 

Inspired by the recent success of deep hashing methods, the objective of this work is to examine the feasibility of integrating deep hashing with a secure architecture to generate a secure multimodal template \commentAQ{for face and iris biometrics.} Contributions include:

\begin{itemize}
    \item We use deep hashing to generate a binary latent shared representation from \commentAQ{a user's face and iris biometrics.}
    \item We combine cancelable biometrics and secure sketch schemes to create a hybrid secure architecture.
    \item We integrate the hybrid secure architecture with the deep hashing framework to generate a multimodal secure sketch, which is cryptographically hashed to generate the secure multimodal template.
    \item We analyze the trade-off between genuine accept rate (GAR) and security for the proposed secure multimodal scheme using an actual multimodal database.
    \item Additionally, we also perform an information-theoretic privacy analysis, and unlinkability analysis for the proposed secure system. 
    \end{itemize}

\commentAQ{The proposed approach represents a biometric security framework integrated with multimodal fusion and deep hashing, and is particularly well suited for structural data from modalities like face and iris. Our approach is not a general biometric security framework that can be applied to all biometric modalities, but rather a proposal that needs further study and validation.}

The rest of the paper is organized as follows. Section \ref{sec:related_work} provides a background on deep hashing techniques and the various multibiometric template security schemes proposed in the literature. The proposed framework and the associated algorithms are introduced in Section \ref{sec:arch}. Implementation details are presented in Section \ref{sec:impl}. In Section \ref{sec:CTM_EXPR}, we present a performance evaluation of the cancelable biometric module, which is a part of the overall proposed system. The performance evaluation of the overall proposed system is discussed in Section \ref{sec:SSTM_EXPR}. The conclusions are summarized in Section \ref{sec:conc}.
\vspace{-0.25cm}
\section{Related Work}\label{sec:related_work}


\subsection{Deep Learning}

Deep learning has emerged as a new area of machine learning and is being extensively applied to solve problems that have resisted the best attempts of the machine learning and artificial intelligence community for many years. It has turned out to be very good at discovering intricate structures in high-dimensional data and is therefore applicable to many domains of science, business, and government.

Deep learning has been extensively implemented and applied to image recognition tasks. Krizhevsky \etal \  \cite{krizhevsky_imagenet_2012} provided a breakthrough in the field of object recognition and ImageNet classification by applying a CNN for object recognition. They were able to reduce the error rate by almost half. The neural network implemented in \cite{krizhevsky_imagenet_2012} is currently known as \emph{AlexNet} and triggered the rapid endorsement of deep learning by the computer vision community. Simonyan \etal \  \cite{simonyan_very_deep_2014} increased the depth of the convolutional network but reduced the size of the filters being used for convolution. The main contribution in \cite{simonyan_very_deep_2014} was a thorough evaluation of networks of increasing depth using an architecture with very small $3 \times 3$ convolution filters, which represented a compelling advancement over the prior-art configurations. 


Szegedy \etal \ \cite{szegedy_googlenet_2014} advanced the architecture of CNN by making it deeper, similar to  \cite{krizhevsky_imagenet_2012}, and wider by introducing a CNN termed \emph{inception}. One particular incarnation of this architecture is known as \emph{GoogleNet} which is 22 layers deep. He \etal \ \cite{he_resnet_2016} developed a very deep 152 layer convolutional neural network architecture named \emph{ResNet}. The novelty of \emph{ResNet} lies not only in creating a very deep network but also in the use of a residual architecture to reformulate the layers as learning residual functions with reference to the layer inputs, instead of learning unreferenced functions. 



In addition to improving performance in image and speech recognition \cite{krizhevsky_imagenet_2012,simonyan_very_deep_2014,he_resnet_2016,farabet_scene_labeling_2012}, deep learning has produced extremely promising results for various tasks in natural language understanding, particularly topic classification, sentiment analysis, question answering, and language translation.

\vspace{-0.30cm}
\subsection{Deep Hashing}
Many hashing methods \cite{iterative_gong_2013,k_means_he_2013,fast_jain_2008,learning_kulis_2009,minimal_norouzi_2011,raginsky_2009_locality} have been proposed to enable efficient approximate nearest neighbor search due to low space and time complexity. These traditional hashing methods can be categorized into data-independent or data-dependent methods. A comprehensive survey of hashing techniques is presented in \cite{wang_2014_hashing}. Initial research on hashing was mainly focused on data-independent methods, such as locality sensitive hashing (LSH). LSH methods \cite{gionis_1999_similarity} generate hashing bits by using random projections. However, LSH methods demand a significant amount of memory as they require long codes to achieve satisfactory performance. 

To learn compact binary codes, data-dependent hashing methods have been proposed in the literature. Data-dependent methods learn similarity-preserving hashing functions from a training set. Data-dependent hashing methods can be categorized as \emph{unsupervised} \cite{weiss_2009_spectral,iterative_gong_2013,smeulders_2000_content} or \emph{supervised} \cite{learning_kulis_2009,minimal_norouzi_2011}. These methods have achieved success to some extent by using handcrafted features for learning hash functions. However, the handcrafted features do not preserve the semantic data similarities of image pairs and non-linear variation in real-world data \cite{deep_hashing_liu_2016}. This has led to a surge of deep hashing methods \cite{supervised_xia_2014,simultaneous_lai_2015,learning_lin_2016,deep_hashing_liu_2016,Yuan_2018_ECCV,Yuan_2018_DeepHV} where deep neural networks encode non-linear hash functions. This leads to an effective end-to-end learning of feature representation and hash coding. 

Xia \etal \  \cite{supervised_xia_2014} adopted a two-stage learning strategy wherein the first stage computes hash codes from the pairwise similarity matrix and the second stage trains a deep neural network to fit the hash codes generated in the first stage. The model proposed by Lai \etal \  \cite{simultaneous_lai_2015} simultaneously captures the intermediate image features and trains the hashing function in a joint learning process. The hash function in \cite{simultaneous_lai_2015} uses a divide-and-encode module, which splits the image features derived from the deep network into multiple blocks, each block encoded into one hash bit. Liu \etal \  \cite{deep_hashing_liu_2016} present a deep hashing model that learns the hash codes by simultaneously optimizing a contrastive loss function for input image pairs and imposing a regularization on the real-valued outputs to approximate the binary values. Zhu \etal \  \cite{zhu_2016_deep} proposed a deep hashing method to learn hash codes by optimizing a pairwise cross-entropy quantization loss to preserve the pairwise similarity and minimize the quantization error simultaneously.
\vspace{-0.32cm}
\subsection{Secure Biometrics}
The leakage of biometric template information to an adversary constitutes a serious threat to security and privacy of the user because if an adversary gains access to the biometric database, he can potentially obtain the stored biometric information of a user. To alleviate the security and privacy concerns in biometric usage, secure biometric architectures have been developed to allow authentication without requiring that the reference biometric template be stored in its raw format at the access control device. Secure biometric architectures include \emph{biometric cryptosystems} (e.g., fuzzy commitment and secure sketch) \cite{Sutcu_2007_Protecting,Juels_1999_Fuzzy,Nandakumar_2007_Fingerprint_FV,Nagar_2008_Securing} and \emph{transformation based methods} (e.g., cancelable biometrics) \cite{rane_secure_biom_2013}. 


Fuzzy commitment, a classical method of biometric protection, was first proposed in 1999 \cite{Juels_1999_Fuzzy}. Forward error correction (FEC) based fuzzy commitment can also be viewed as a method of extracting a secret code by means of polynomial interpolation \cite{Juels_2002_Vault}. An implementation example of such a fuzzy commitment scheme appears in \cite{Nagar_2008_Securing}, wherein a BCH code is employed for polynomial interpolation; experiments show that when the degree of the interpolated polynomial is increased, the matching becomes more stringent, reducing the false accept rate (FAR), but increasing the false reject rate (FRR).


Cancelable biometrics was first proposed by Ratha \etal \  \cite {ratha_canc_biom_2007}, after which, there have been various different methods of generating cancelable biometric templates. Some popular methods use non-invertible transforms \cite {ratha_canc_biom_2007}, bio-hashing \cite{kong_analysis_2006}, salting \cite{zuo_cancelable_2008} and random projections \cite {teoh_cancellable_2008}. Literature surveys on cancelable biometrics can be found in \cite {rane_secure_biom_2013}, and \cite{patel_canc_biom_2015}.  

\vspace{-0.35cm}

\subsection{Secure Multimodal Biometrics}

The secure biometric frameworks have been extended to include multiple biometric traits of a user \cite{sutcu_secure_2007,nagar_multibiometriccryptosystems_2012,nandakumar_multibiometric_2008,canuto_investigating_2013}. In \cite { sutcu_secure_2007} face and fingerprint templates are concatenated to form a single binary string and this concatenated string is used as input to a secure sketch scheme. Kelkboom \etal \  \cite{kelkboom_multi_algorithm_2009} provided results for decision-level, feature-level, and score-level fusion of templates by using the number of errors corrected in a biometric cryptosystem as a measure of the matching score. 

Nagar \etal \ \cite{nagar_multibiometriccryptosystems_2012} developed a multimodal cryptosystem based on feature-level fusion using two different security architectures, \emph{fuzzy commitment} and \emph{fuzzy vault}. Fu \etal \  \cite{fu_multibiometric_2009} theoretically analyzed four different versions of the multibiometric cryptosystem: \emph{no-split}, \emph{MN-split}, \emph{package}, and \emph{biometric model}, using template security and recognition accuracy as performance metrics. In the first three versions, the biometric templates are secured individually with a decision-level fusion, while the last version is a feature-level fusion.

Research has also been directed towards integrating cancelable biometric techniques into multimodal systems. Canuto \etal \  \cite{canuto_investigating_2013} combined voice and iris using cancelable transformations and decision level fusion. Paul and Gavrilova \cite{paul_multimodal_2012} used random projections and transformation-based feature extraction and selection to generate cancelable biometric templates for face and ear. There are some studies related to the use of multi-feature biometric fusion, which involves combining different features of the same biometric trait \cite{rathgeb_cancelable_2014}. 

However, none of the above papers present a secure architecture that combines multiple secure schemes to protect multiple biometrics of a user. In this paper, we have integrated a deep hashing framework with a hybrid secure architecture by combining cancelable biometric templates and secure sketch, which makes it computationally prohibitive to forge a combination of multiple biometrics that passes the authentication. 



\vspace{-0.25cm}
\section{Proposed Secure Multibiometric System}\label{sec:arch}
\subsection{System Overview}
\begin{figure}[t]
\centering
\includegraphics[width=8.0cm]{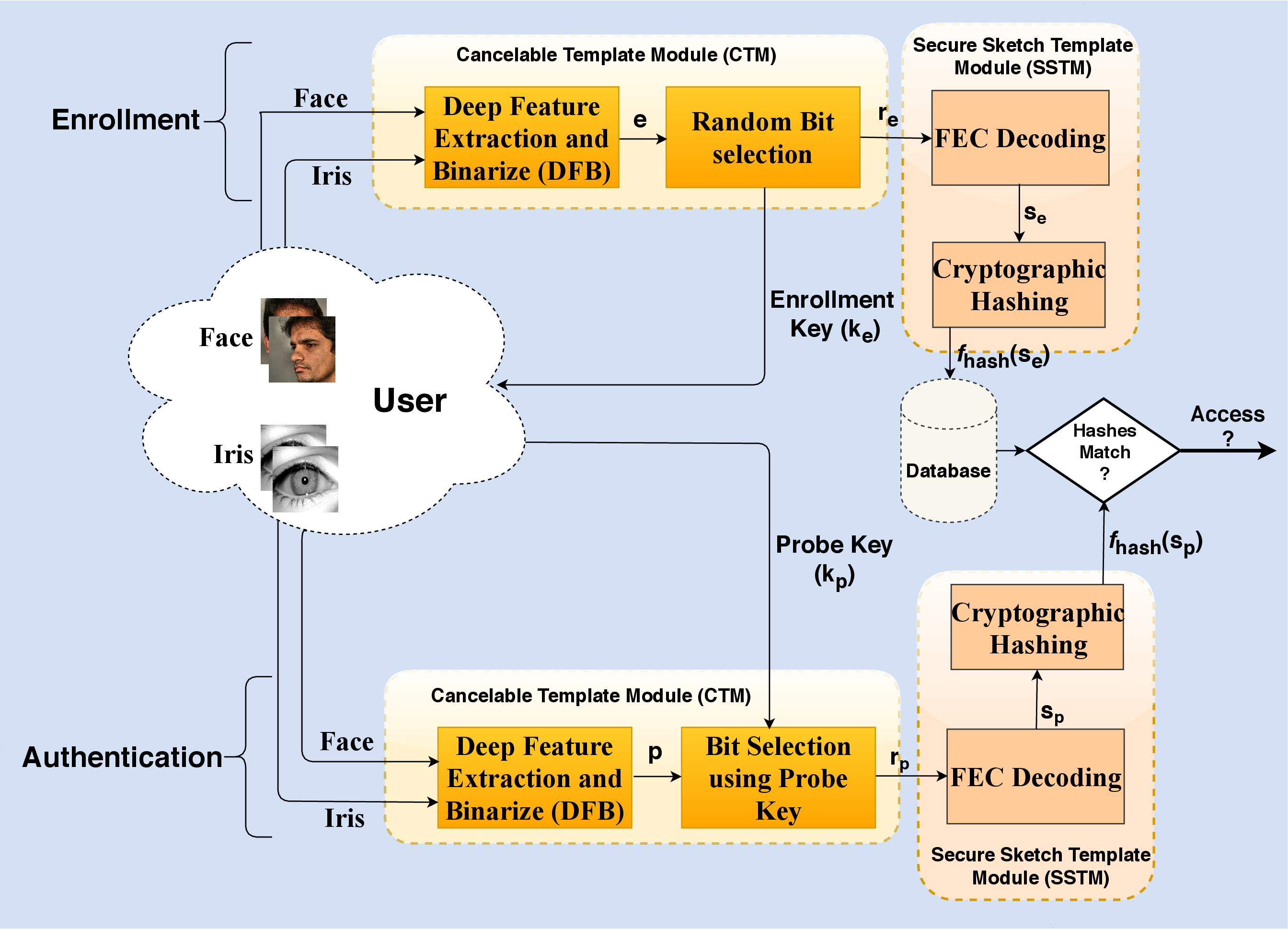}
\caption{Block diagram of the proposed system.}\label{fig:enrol} 
\vspace{-0.35cm}
\end{figure}

In this section, we present a system overview including descriptions of the enrollment and authentication procedures. We propose a feature-level fusion  and hashing framework for the secure multibiometric system. The general framework for the proposed secure multibiometric system is shown in Fig. \ref{fig:enrol}. During enrollment, the user provides their biometrics (e.g., face and iris) as an input to the deep feature extraction and binarization (DFB) block. The output of the DFB block is an $J$-dimensional binarized joint feature vector \textbf{e}. A random selection of feature components (bits) from the binarized joint feature vector \textbf{e} is performed. The number of random components that are selected from the binarized joint feature vector \textbf{e} is $G$. The indices of these randomly selected $G$ components forms the enrollment key $\textbf{k}_\textbf{e}$, which is given to the user. The cancelable multimodal template $\textbf{r}_\textbf{e}$ is formed by selecting the values from the vector \textbf{e} at the corresponding location or indices as specified by the user-specific key $\textbf{k}_\textbf{e}$. 


This random selection of $G$ components from the binarized joint feature vector \textbf{e} helps in achieving revocability, because if a key is compromised, a new key can be issued with a different set of random indices. In the next step, $\textbf{r}_\textbf{e}$ is passed through a forward error correction (FEC) decoder to generate the multimodal sketch $\textbf{s}_\textbf{e}$. The cryptographic hash of this sketch $f_\mathsf{hash}$($\textbf{s}_\textbf{e}$) is stored as a secure template in the database.   

During authentication, the probe user presents the biometrics and the key $\textbf{k}_\textbf{p}$ where $\textbf{k}_\textbf{p}$ could be same as the enrollment key $\textbf{k}_\textbf{e}$ in the case of a genuine probe or it could be a synthesized key in case of an impostor probe. The probe biometrics are passed through the DFB block to obtain a binary vector \textbf{p}, which is the joint feature vector corresponding to the probe. Using the key $\textbf{k}_\textbf{p}$ provided by the user, the multimodal probe template $\textbf{r}_\textbf{p}$  is generated by selecting the values from \textbf{p}  at the locations given by the key $\textbf{k}_\textbf{p}$. In the next step, $\textbf{r}_\textbf{p}$ is passed through a FEC decoder with the same code used during enrollment to generate the probe multimodal sketch $\textbf{s}_\textbf{p}$. If the cryptographic hash of the enrolled sketch $f_\mathsf{hash}$(\textbf{s}$_\textbf{e}$) matches the cryptographic hash of the probe sketch $f_\mathsf{hash}$(\textbf{s}$_\textbf{p}$), then the access is granted, otherwise the access is denied.

The proposed secure multibiometric system consists of two basic modules: \emph{Cancelable Template Module (CTM)} and \emph{ Secure Sketch Template  Module (SSTM)}, which are described more fully in the following subsections.


\vspace{-0.35cm}
\subsection{Cancelable Template Module }\label{subsec:CTM}
\begin{figure}[t]
\vspace{-0.20cm}
\centering
\includegraphics[width=8.0cm]{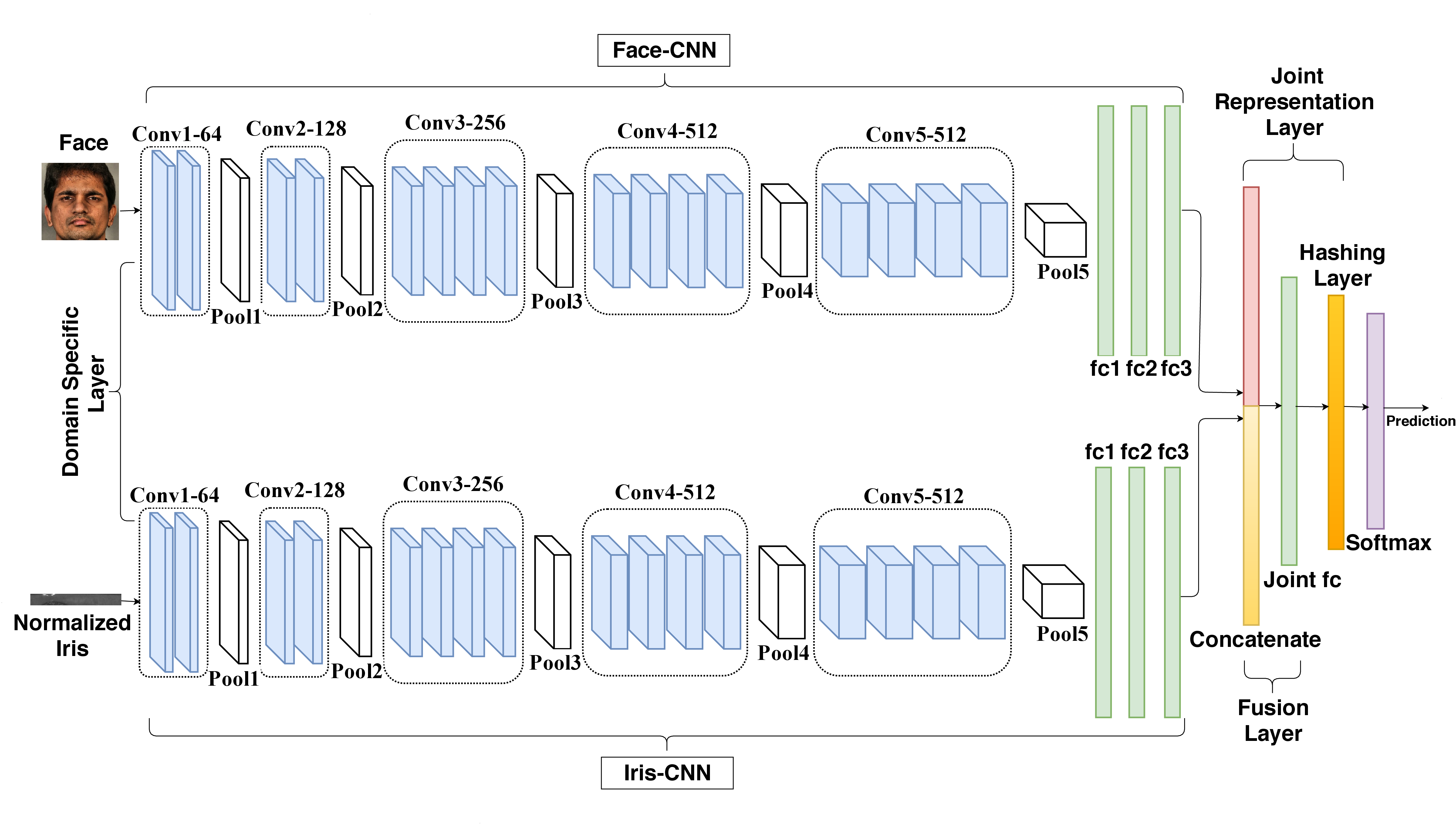}
\caption{Proposed deep feature extraction and binarization (DFB) model for the fully concatenated architecture (FCA).}\label{fig:arch_fca} 
\vspace{-0.45cm}
\end{figure}

The cancelable template module (CTM) consists of two blocks: DFB block and random-bit selection block. The primary function of CTM is non-linear feature extraction, fusion, and binarization using the proposed DFB architecture shown in Figs. \ref{fig:arch_fca} and \ref{fig:arch_bla}. The DFB consists of two layers: domain-specific layer (DSL) and joint representation layer (JRL). 

\subsubsection{\textbf{Domain-Specific Layer}}\label{subsubsec:dsl}
The DSL consists of a CNN for encoding the face (``Face-CNN") and a CNN for encoding the iris (``Iris-CNN"). For each CNN, we use VGG-19 \cite{simonyan_very_deep_2014} pre-trained on ImageNet \cite{ILSVRC15} as a starting point and then fine-tune it with an additional fully connected layer \emph{fc3} as described in Sec. \ref{subsec:nwparam_face} and \ref{subsec:nwparam_iris}. \comment{ There are multiple reasons for using VGG-19 pre-trained on the ImageNet dataset for encoding the face and iris. In the proposed method, the VGG-19 is only used as feature-extractor for face and iris modalities. It can be seen from the previous literature \cite{Nguyen_2017_IrisRW,zhao_2017_towards,minaee_2016_experimental,Schroff_2015_FaceNetAU,Sun_2015_DeepID3,Parkhi_2015_DeepFR} that the features provided by a VGG-19 pre-trained on ImageNet and fine-tuned on face/iris images are very discriminative and therefore can be used for face/iris recognition. Moreover, starting with a well-known architecture and using the same architecture for both modalities makes the work highly reproducible.}

 

\subsubsection{\textbf{Joint Representation Layer}}\label{subsubsec:jrl}

The output feature vectors of the Face-CNN and Iris-CNN are fused and binarized in the JRL, which is split into two sub-layers: fusion layer and hashing layer. The main function of the fusion layer is to fuse the individual face and iris representations from domain-specific layers into a shared multimodal feature embedding. The hashing layer binarizes the shared multimodal feature representation that is generated by the fusion layer.

\begin{figure}[t]
\centering
\includegraphics[width=8.0cm]{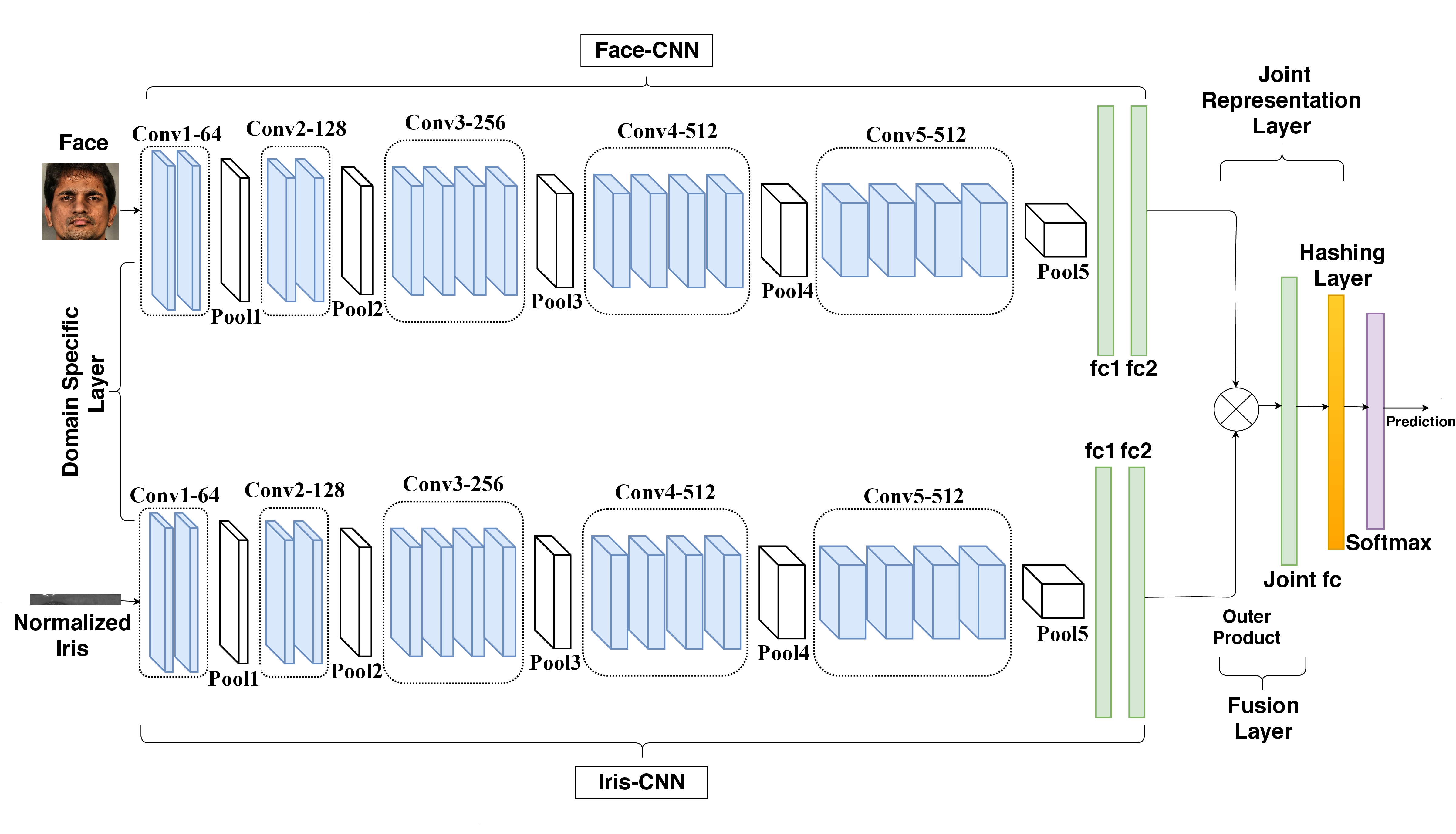}
\caption{Proposed deep feature extraction and binarization (DFB) model for the bilinear architecture (BLA).}\label{fig:arch_bla} 
\vspace{-0.45cm}
\end{figure}



\textbf{Fusion layer}: We have implemented two different architectures for the fusion layer: (1) fully concatenated architecture (FCA), and (2) bilinear architecture (BLA). These two architectures differ in the way the face and iris feature vectors are fused together to generate the joint feature vector.

In the FCA shown in Fig. \ref{fig:arch_fca}, the outputs of the Face-CNN and Iris-CNN are concatenated vertically using a concatenation layer. The concatenated feature vector is passed through a fully connected layer (hereon known as \emph{joint fully connected layer}) which reduces the feature dimensionality (i.e., the number of dimensions is reduced) and also fuses the iris and face features. In the FCA, the concatenation layer and the joint fully connected layer together constitute the fusion layer. 

In the BLA shown in Fig. \ref{fig:arch_bla}, the outputs of the Face-CNN and Iris-CNN are combined using the matrix outer product; i.e., the bilinear feature combination of column face feature vector $\textbf{f}_{\mathsf{face}}$ and column iris feature vector $\textbf{f}_{\mathsf{iris}}$ given by  $\textbf{f}_{\mathsf{face}}\textbf{f}_{\mathsf{iris}}^{T}$. Similar to the FCA, the bilinear feature vector is also passed through a joint fully connected layer. In the BLA, the outer product layer and the joint fully connected layer together constitute the fusion layer. 

\comment{In addition to the two techniques (FCA, BLA) used in this paper, there could be other fusion techniques for combining multiple modalities \cite{feichtenhofer_2016_convolutional}. The rationale behind implementing FCA is that we wanted to use a fusion technique that involves just simple concatenation where there is no interaction between the two modalities being fused before the joint fully connected layer (Joint $f_c$). As evident from Fig. \ref{fig:arch_fca}, the iris and face extracted features do not interact with each other and have their own network parameters before passing through the joint fully connected layer. On the other hand, we also wanted to test a fusion technique that involves high interactions between the two modalities feature vectors at every element before being passed through the joint fully connected layer. That is the reason we have used BLA, which is based on \emph{bilinear fusion} \cite{lin_2015_bilinear}. Bilinear fusion exploits the higher-level dependencies of the modalities being combined by considering the pairwise multiplicative interactions between the modalities at each feature element (i.e., matrix outer product of modalities feature vector). Moreover, bilinear fusion is widely being used in many CNN applications such as fine-grained visual recognition and video action recognition \cite{lin_2015_bilinear,feichtenhofer_2016_convolutional}.}

\textbf{Hashing layer}: The output of the fusion layer produces a $J$-dimensional shared multimodal feature vector of real values. We can directly binarize the output of the fusion layer by thresholding at any numerical value or at the population mean. However, this kind of thresholding leads to a quantization loss, which results in sub-optimal binary codes. To account for this quantization loss, we have included another latent layer after the fusion layer, which is known as the hashing layer (shown in orange in Fig. \ref{fig:arch_fca} and \ref{fig:arch_bla}). The main function of the hashing layer is to binarize (hash) the shared multimodal feature representation generated by the fusion layer.  
 
One key challenge of implementing deep learning to hash end-to-end is converting deep representations, which are real-valued and continuous, to exactly binary codes. The sign activation function $h=\mbox{sgn}(z)$ can be used by the hashing layer to generate the binary hash codes. However, the use of the non-smooth sign-activation function makes standard back-propagation impracticable as the gradient of the sign function is zero for all non zero inputs. The problem of zero gradient at the hashing layer due to a non-smooth sign activation can be diminished by using the idea of continuation methods \cite{cao_2017_hashnet}. 

 

 We circumvent the zero-gradient problem by starting with a smooth activation function $y=\mbox{tanh}(\beta x)$ and making it sharper by increasing the bandwidth $\beta$ as the training proceeds. We have utilized a key relationship between the sign activation function and the scaled $\mbox{tanh}$ function using limits: \begin{equation}\lim_{\beta\to\infty} \mbox{tanh}(\beta x )=\mbox{sgn}(x),\label{eq:1}\end{equation} where $\beta >0$ is a scaling parameter. The scaled function $\mbox{tanh}(\beta x)$ will become sharper and more saturated as we increase $\beta$ during training. Eventually, this non-smooth $\mbox{tanh}$ function with $\beta\to\infty$ converges to the original, difficult to optimize, sign activation function. For training the network, we start with a $\mbox{tanh}(\beta x)$ activation for the hashing layer with $\beta=1$ and continue training until the network converges to zero loss. We then increase the value of $\beta$ while holding other training parameters equal to the previously converged network parameters, and start retraining the network for convergence. This process is repeated several times by increasing the bandwidth of the $\mbox{tanh}$ activation as $\beta\to\infty$ until the hashing layer can generate binary codes. In addition to using this continuation method for training the network, we have used additional cost functions for efficient binary codes. The overall objection function used for training is discussed in Sec. \ref{subsec:obj}

\subsubsection{\textbf{Random-Bit Selection}}

One of the most prevalent methods for generating cancelable template involves random projections of the biometric feature vector \cite{teoh_cancellable_2008}, in which the random projection is a revocable transformation. Similarly, the DFB architecture is considered to be the projection of the biometric images in a $J$-dimensional space. The randomness and revocability is added by performing a random bit selection of $G$ bits from the $J$-dimensional output vector $\textbf{e}$ of the DFB. After the selection, these random bits are then arranged in descending order of reliability. The reliability of each bit is computed as $((1-p^{e}_g)p^{e}_i)$, where $p^{e}_i$ and $p^{e}_g$ are the impostor and genuine bit error probabilities, respectively\cite{nagar_multibiometriccryptosystems_2012}. A different set of random bits is selected for every user and these randomly selected $G$ bits form the cancelable multimodal template $\textbf{r}_\textbf{e}$ and the indices of the selected bits forms the key for that user $\textbf{k}_\textbf{e}$. This key is revocable and a new set of random bits can be selected in case the key gets compromised. \comment{Selecting a new set of bits requires that either the original vector \textbf{e} be retrieved from a secure location or else the user is re-enrolled, thereby presenting a new instance of \textbf{e}}. This method of using the DFB architecture with a random bit selection is analogous to a random projection as a revocable transformation to generate a cancelable template \cite{teoh_cancellable_2008}. 

\comment{It is important to note that even if multiple users end up having the same key $\textbf{k}_\textbf{e}$ (i.e., same indices of $G$ random bits), their final templates will still be distinct because the template depends on the values at those $G$ bits (i.e., $\textbf{r}_\textbf{e}$) from the enrollment vector \textbf{e}, and not only on the indices of the $G$ bits. A second user having the same key $\textbf{k}_\textbf{e}$ is equivalent to the stolen key scenario, which is analyzed in Sec. \ref{subsec:CTMPEEV}}. 

\vspace{-0.33cm}
\subsection{Secure Sketch Template Module}\label{subsec:SSTB}

As shown in Fig. \ref{fig:enrol}, the cancelable template (output of CTM) $\textbf{r}_\textbf{e}$ is an intermediate template and is not stored in the database. The cancelable template is passed through the SSTM to generate the secure multimodal template, which is stored in the database. As the name suggests, the SSTM module is related to the secure sketch biometric template protection scheme. The SSTM contains two important blocks: FEC decoding and cryptographic hashing. The main function of the SSTM is to generate a multimodal secure sketch by using the cancelable template as an input to the FEC decoder. This multimodal secure sketch (output of the FEC decoder) is cryptographically hashed to generate the secure multimodal template, which is stored in the database.  

The FEC decoding implemented in our framework is the equivalent of a secure-sketch template protection scheme. In a secure-sketch scheme, sketch or helper data is generated from the user's biometrics and this sketch is stored in the access-control database. A common method of implementing secure sketch is to use error control coding. In this method error control coding is applied to the biometrics or the feature vector to generate a sketch which is stored in the database. Similarly, in our framework, the FEC decoding is considered to be the error control coding part required to generate the secure sketch. Our approach is different from other secure sketch approaches using error correcting codes (ECC) as we do not have to present any other side information to the decoder like a syndrome or a saved message key \cite{Sutcu_2008_Feature_SW}. 

The cancelable template $\textbf{r}_\textbf{e}$ from the CTM is considered to be the noisy codeword of an ECC that we can select. This noisy codeword is decoded with a FEC decoder and the output of the decoder is the multimodal secure sketch $\textbf{s}_\textbf{e}$ that corresponds to the codeword closest to the cancelable template. This multimodal sketch $\textbf{s}_\textbf{e}$ is cryptographically hashed to generate $f_\mathsf{hash}$($\textbf{s}_\textbf{e}$) stored in the database. 

During authentication, the same process is performed. The probe user provides the biometrics and the key which are used to generate the probe template $\textbf{r}_\textbf{p}$. The probe template $\textbf{r}_\textbf{p}$ is passed through an FEC decoder for the same error correcting code used during the enrollment. The output of the FEC decoder is the probe multimodal sketch $\textbf{s}_\textbf{p}$ which is cryptographically hashed and access is granted only if this hash matches the enrolled hash. During authentication, if it is a genuine probe, the enrollment $\textbf{r}_\textbf{e}$ and the probe vector $\textbf{r}_\textbf{p}$ would usually decode to the same codeword in which case the hashes would match and access would be granted.

\section{Implementation} \label{sec:impl}
\subsection{Objective Function for Training the Deep Hashing Network} \label{subsec:obj}

In this section, the objective function used for training the deep hashing network is described. 

\textbf{Semantics-preserving binary codes}: In order to construct semantics-preserving binary codes, we propose to model the relationship between the labels and the binary codes. Every input image is associated with a semantic label, which is derived from the hashing layer's binary-valued outputs, and the classification of each image is dependent on these binary outputs. Consequently, we can ensure that semantically similar images belonging to the same subject are mapped to similar binary codes through an optimization of a loss function defined on the classification error. The classification formulation has been incorporated into the deep hashing framework by adding the \emph{softmax} layer as shown in Fig. \ref{fig:arch_fca} and Fig. \ref{fig:arch_bla}. Let $E_{1}$ denote the objective function required for classification formulation: 
\vspace{-0.20cm}
\begin{equation} E_{1}(\textbf{w})=\frac{1}{N}\sum_{n=1}^{N}L_n(f(x_{n},\textbf{w}),y_{n}) + \lambda ||\textbf{w}||^{2} \label{eq:2},\end{equation} where the first term $L_n(.)$ is the classification loss for a training instance $n$ and is described below, $N$ is the number of training images in a mini-batch. $f(x_{n},\textbf{w})$ is the predicted softmax output of the network and is a function of the input training image $x_n$ and the weights of the network \textbf{w}. The second term is the regularization function where $\lambda$ governs the relative importance of the regularization. 

The choice of the loss function $L_n(.)$ depends on the application itself. We use a classification loss function that uses softmax outputs by minimizing the cross-entropy error function. Let the predicted softmax output $f(x_{n},\textbf{w})$ be denoted by $\hat{y}_{n}$. The classification loss for the $n^\mathsf{th}$ training instance is: 

\vspace{-0.20cm}
\begin{equation}L_{n}(\hat{y}_{n},y_{n})=-\sum_{m=1}^{M}y_{n,m}\ln \hat{y}_{n,m} \label{eq:3},\end{equation} where $y_{n,m}$ and $\hat{y}_{n,m}$ is the ground truth and the prediction result for the $m^\mathsf{th}$ unit of the $n^\mathsf{th}$ training instance, respectively and $M$ is the number of output units.

\textbf{Additional cost constraints for efficient binary codes}: The continuation method that has been described in \ref{subsubsec:jrl} forces the activations of the hashing layer closer to -1 and 1. However, we need to include additional cost constraints to obtain more efficient binary codes. 

Let the $J$-dimensional vector output of the hashing layer be denoted by $\textbf{o}^{H}_{n}$ for the $n$-th input image, and let the $i$-th element of this vector be denoted by $o^{H}_{n,i} (i=1,2,3, \cdots,J)$. The value of $o^{H}_{n,i}$ is in the range of $[-1,1]$ because it has been activated by the $\mbox{tanh}$ activation. To make the codes closer to either -1 or 1, we add a constraint of maximizing the sum of squared errors between the hashing layer activations and 0, which is given by $\sum_{n=1}^{N}||\textbf{o}^{H}_{n}-\textbf{0}||^{2}$, where $N$ is the number of training images in a mini-batch and \textbf{0} is the $J$-dimensional vector with all elements equal to 0. However, this is equivalent to maximizing the square of the length of the vector formed by the hashing layer activations, that is $\sum_{n=1}^{N}||\textbf{o}^{H}_{n}-\textbf{0}||^{2}=\sum_{n=1}^{N}||\textbf{o}^{H}_{n}||^{2}$. Let $E_{2}(\textbf{w})$ denote this constraint to boost the activations of units in hashing layer to be closer to -1 or 1:
\vspace{-0.20cm}
\begin{equation}E_{2}(\textbf{w})=-\frac{1}{J}\sum_{n=1}^{N}||\textbf{o}^{H}_{n}||^{2}  \label{eq:4}.\end{equation}


In addition to forcing the codes to become binarized, we also require that the codes satisfy a balance property whereby they produce an equal number of -1's and 1's, which maximizes the entropy of the discrete distribution and results in hash codes with better discrimination. To achieve the balance property, we want each bit to fire  $50\%$ of the time by minimizing the sum of the squared error between the mean of the hashing layer activations and $0$. This is given by $\sum_{n=1}^{N}(\text{mean}(\textbf{o}^{H}_{n})-0)^{2}$, which is equivalent to $\sum_{n=1}^{N}(\text{mean}(\textbf{o}^{H}_{n}))^{2}$ where mean(.) computes the average of the elements of the vector. This criterion helps to obtain binary codes with an equal number of -1's and 1's. Let $E_{3}(\textbf{w})$ denote this constraint that forces the output of each node to have a $50\%$ chance of being -1 or 1:
\vspace{-0.15cm}
\begin{equation}E_{3}(\textbf{w})= \sum_{n=1}^{N}(\text{mean}(\textbf{o}^{H}_{n}))^{2}\label{eq:5}.\end{equation}


Combining the above two constraints (binarizing and balance property constraints) makes $\textbf{o}^{H}_{n}$ close to a length $J$ binary string with a $50\%$ chance of each bit being -1 or 1.



\textbf{Overall objective function}: The overall objective function to be minimized for a semantics-preserving efficient binary codes is given as:
\vspace{-0.20cm}
\begin{equation}\alpha E_{1}(\textbf{w})+ \beta E_{2}(\textbf{w}) + \gamma E_{3}(\textbf{w}) \label{eq:6},\end{equation} where $\alpha$, $\beta$, and $\gamma$ are the tuning parameters of each term. The optimization to be performed to minimize the overall objective function is given as:
\vspace{-0.20cm}
\begin{equation}\textbf{w}=\operatorname*{arg\,min}_\textbf{w} (\alpha E_{1}(\textbf{w})+ \beta E_{2}(\textbf{w}) + \gamma E_{3}(\textbf{w}))\label{eq:6b} \vspace{-0.20cm} \end{equation}





The optimization given in (\ref{eq:6b}) is the sum of the losses form and can be performed via the stochastic gradient descent (SGD) efficiently by dividing the training samples into batches. For training the JRL we adopt a two-step training procedure where we first train only the JRL using the objective function in (\ref{eq:6}) greedily with softmax by freezing the Face-CNN and Iris-CNN. After training the  JRL, the entire model is fine-tuned end-to-end using the same objective function with back-propagation at a relatively small learning rate. 



\comment{
For tuning the hyper-parameters $\alpha$, $\beta$, and $\gamma$ of the objective function (\ref{eq:6}), we have utilized an iterative grid search. To start, consider a cubic grid with all possible values for each parameter. Each point on this grid ($\alpha$,$\beta$,$\gamma$) represents a combination of the three hyper-parameters. Because exhaustively searching over all combinations is computationally expensive, we adopted an iterative and adaptive grid search.}

\comment{In the iterative and adaptive grid search, for each hyper-parameter, we considered the set of values $\mathcal{S} = \{ 1, 2i \}$ for $i=\{1, ..., 15\}$; i.e., the set containing 1 and all positive even integers from 2 to 30. This grid search is performed iteratively, where each iteration is a combination of 3 steps.
In the first step, we fixed $\alpha$, and $\gamma$ to be 1 and $\beta$ is chosen from the set $\mathcal{S}$. Therefore the set of points considered for this step is:
\vspace{-0.20cm}
\begin{equation}
    (\alpha,\beta,\gamma)=(1,\beta_i,1), \text{where}    \ \beta_i\in\mathcal{S}.
\end{equation}
For each point in the above set $(1,\beta_{i},1)$, we trained our DFB network and calculated the genuine accept rate (GAR) for the overall system for a security of 104 bits using a 5-fold cross validation. Using this method, we found the best value for hyper-parameter $\beta$ that gave us the highest GAR with the values of $\alpha$ and $\gamma$ as 1. This best value of $\beta$ will be denoted as $\beta^t$ where the superscript $t$ signifies the iteration number.} 

\comment{In the second step, we repeated the same process with $\alpha$ and $\beta$ fixed at 1 and choosing $\gamma$ from the set $\mathcal{S}$:
\begin{equation}
    (\alpha,\beta,\gamma)=(1,1,\gamma_i), \text{where}   \ \gamma_i\in\mathcal{S}.
\end{equation}
Again using a 5-fold cross validation, we found the best value for hyper-parameter $\gamma$, which is denoted by $\gamma^1$, that gave us the highest GAR with the values of $\alpha$ and $\beta$ fixed as 1. In the third step, the same procedure was performed by keeping $\beta$, and $\gamma$ fixed at 1 and found the best value for hyper-parameter $\alpha$, which is denoted by $\alpha^1$, from the set $\mathcal{S}$.
These three steps together complete one iteration of the iterative grid search.}

\comment{In the next iteration, we again performed the above 3 steps but instead of fixing the values of the two parameters to 1, we fixed the value of the two parameters to be the best value found in the previous iteration for those parameters. To explain this, consider the best value of the 3 parameters found in the first iteration, denoted by $\alpha^1$,$\beta^1$,$\gamma^1$. In the first step of the second iteration, we fixed $\alpha$, and $\gamma$ to be $\alpha^1$ and $\gamma^1$ respectively and chose $\beta$ from the set $\mathcal{S}$. Therefore the set of points are:
\vspace{-0.15cm}
\begin{equation}
    (\alpha,\beta,\gamma)=(\alpha^1,\beta_i,\gamma^1), \text{where}   \ \beta_i\in\mathcal{S}.
\end{equation}
Again, using a 5-fold cross validation, we found the best value for hyper-parameter $\beta$ with the other parameters set to $\alpha^1$ and $\gamma^1$. This best value of $\beta$ will be denoted as $\beta^2$ since this is the second iteration. Similarly, we performed the second and third steps of the second iteration to find the $\gamma^2$ and $\alpha^2$ respectively.}

\comment{ We continued performing these iterations until the parameters converged, which implies that the best value of each parameter did not change from one iteration to the other; i.e., $\alpha^t=\alpha^{t-1}, \beta^t=\beta^{t-1}, \gamma^t=\gamma^{t-1}.
$} 

\comment{Using the above procedure for hyperparameter tuning, we have found the values of $\alpha^t$, $\beta^t$, and $\gamma^t$ to be 8, 2, 2 for FCA and 6, 4, 2 for BLA respectively. The importance of each term will be further discussed in the ablation study in Section \ref{subsec:hyper_tun}.}
\vspace{-0.35cm}
\subsection{Network parameters for the Face-CNN}\label{subsec:nwparam_face}

The network used for the Face-CNN is the VGG-19 with an added fully connected layer \emph{fc3} (shown in Fig. \ref{fig:arch_fca}). The Face-CNN is fine-tuned end-to-end with the CASIA-Webface \cite{yi_learning_2014}, which contains 494,414 facial images corresponding to 10,575 subjects. After fine-tuning with CASIA-Webface, the Face-CNN is next fine-tuned with the 2013 session of the WVU-Multimodal face 2012-21013 dataset \cite{wvu_multimodal_2017}. The WVU-Multimodal face dataset for the year 2012 and 2013 together contain a total of 119,700 facial images corresponding to 2263 subjects with 294 common subjects. All the raw facial images are first aligned in 2-D and reduced to a fixed size of
$224\times 224$ before passing through the network \cite{dlib_09}. The only other pre-processing is subtracting the mean RGB value, computed on the training set, from each pixel. The training is carried out by optimizing the multinomial logistic regression objective using mini-batch gradient descent with momentum. The batch size was set to 40, and the momentum to 0.9. The training was regularized by weight decay (the L2 penalty multiplier set to 0.0005) and dropout regularization for the first three fully-connected layers (dropout ratio set to 0.5). We used batch normalization for fast convergence. The learning rate was initially set to 0.1, and then decreased to $90\%$ of its value every 10 epochs. The number of nodes in the last fully connected layer \emph{fc3} before the softmax layer is 1024 for the FCA and 64 for the BLA. This implies that the feature vector extracted from Face-CNN and fused with the feature vector from Iris-CNN has 1024 dimensions for the FCA and 64 for the BLA.
\vspace{-0.35cm}
\subsection{Network parameters for the Iris-CNN}\label{subsec:nwparam_iris}

The network used for the Iris-CNN is the VGG-19 with an added fully connected layer \emph{fc3}. First, the Iris-CNN has been fine-tuned end-to-end using the combination of CASIA-Iris-Thousand \cite{biometrics_ideal_test_2017} and ND-Iris-0405 \cite{bowyer_ndiris_2010} with about 84,000 iris images corresponding to 1355 subjects. Next, the Iris-CNN is fine-tuned using the 2013 session of the WVU-Multimodal iris 2012-21013 dataset \cite{wvu_multimodal_2017}. The WVU-Multimodal iris dataset for the year 2012 and 2013 together contain a total of 257,800 iris images corresponding to 2263 subjects with 294 common subjects. All the raw iris images are segmented and normalized to a fixed size of $64 \times 512$ using Osiris (Open Source for IRIS) which is an open source iris recognition system developed in the framework of the BioSecure project \cite{sutra_2013_biometric}. There is no other pre-processing for the iris images. The other hyper-parameters are consistent with the fine-tuning of the Face-CNN. The iris network has an output of 1024 for FCA and 64 for BLA.
\vspace{-0.35cm}
\subsection{Network parameters for the Joint Representation Layer}

 The details of the network parameters for the two JRL architectures are discussed in this subsection:

\subsubsection{Fully Concatenated Architecture}

In the FCA, the 1024-dimensional outputs of the Face-CNN and Iris-CNN are concatenated vertically to give a 2048-dimensional vector. The concatenated feature vector is then passed through a fully connected layer which reduces the feature dimensionality from 2048 to 1024 and also fuses the iris and face features. The hashing layer is also a fully connected layer that outputs a 1024-dimensional vector and includes a $\mbox{tanh}$ activation. 

For the training of the DFB model, we have used a two-step training procedure. First, only the JRL was trained for 65 epochs, a batch size of 32. The learning rate initially set to 0.1, and then decreased to $90\%$ of its value every 20 epochs. The other hyperparameters are consistent with the fine-tuning of the Face-CNN. After training of the joint representation layer, the entire DFB model was fine-tuned end-to-end for 25 epochs on a batch size of 32. The learning rate initialized to 0.07 which is the final learning rate in the training process of the joint fully connected layer in the first step. The learning rate was decreased to $90\%$ of its value every 5 epochs. For this two-step training process, we have used the 2013 session of the overlap subjects in the 2012 and 2013 sessions from the WVU-Multimodal dataset. This common subset consists of $294$ subjects with a total of $18700$ face and $18700$ iris images with the same number of face and iris images per subject.



\subsubsection{Bilinear architecture}
For the BLA, we do not add \emph{fc3} (i.e., the additional fully connected layer) to either the Face-CNN or the Iris-CNN. In addition, the number of nodes in the first and second fully connected layers $fc1$ and $fc2$ are reduced to 512 and 64, respectively. This means that the output feature vector of the face and iris networks have 64 dimensions rather than the 1024 dimensions of the FCA. The 64-dimensional outputs of the Face-CNN and Iris-CNN are combined in the bilinear (outer product) layer using the matrix outer product as explained in  Sec. \ref{subsubsec:jrl}. The bilinear layer produces an output of dimension $64 \times 64=4096$  fusing the iris and face features. The bilinear feature vector is then passed through a fully connected layer, which reduces the feature dimension from 4096 to 1024 followed by a hashing layer which produces a binary output of 1024 dimensions. 

In the first step of the two-step training process, only the joint representation layer was trained for 80 epochs on a batch size of 32. The momentum was set to 0.9. The learning rate was initially set to 0.1, and then decreased by a factor of 0.1 every two epochs. The other hyperparameters and the input image sizes are consistent with the training process used in FCA. After training of the joint representation layer, the entire DFB model was fine-tuned for 30 epochs on a batch size of 32. The learning rate was initialized to 0.0015 which is the final learning rate in the training process of the joint representation layer in the first step. The learning rate was decreased by a factor of 0.1 every five epochs. The other hyper-parameters are consistent with the training of the JRL in FCA.
\vspace{-0.35cm}
\subsection {Parameters for the FEC Decoding}\label{subsec:exptsetup}

The cancelable template generated from the CTM is considered to be the noisy codeword of some error correcting code that we can select. Due to its maximum distance seperable (MDS) property, we have selected Reed-Solomon (RS) codes and used RS decoder for FEC decoding in SSTM. The $G$-dimensional cancelable template is passed through a Reed-Solomon (RS) decoder to identify the closest codeword, which is the multimodal secure sketch. 

RS codes use symbols of length $m$ bits. The input to the RS decoder is of length $N'= 2^{m-1}$ in symbols, which means the number of bits per input codeword to the decoder is $n'=mN'$. For example, if the symbol size $m=6$ then $N'=63$ is the codeword length in symbols and $n'=378$ is the codeword length in bits. Let's assume the size of the cancelable template is $G=378$ bits, which is the number of bits at the input to the RS decoder. This 378-dimensional vector is decoded to generate a secure sketch whose length is $K'$ symbols or, equivalently, $k'=mK'$ bits. $K'$ can be varied depending on the error correcting capability required for the code and $k'$ also signifies the security of the system in bits \cite{talreja_multibiometric}.

We have used \emph{shortened} RS codes. A shortened RS code is one in which the codeword length is less than $2^{m-1}$ symbols. In standard error control  coding, the shortening of the RS code is achieved by setting a number of data symbols to zero at the encoder, not transmitting them, and then re-inserting them at the decoder. A shortened $[N,K]$ Reed-Solomon code essentially uses an $[N',K']$ encoder, where $N'=2^m-1$, where $m$ is the number of bits per symbol (symbol size) and $K'=K+(N'-N)$. In our experiments we have used $m=8$ and $N'=255$.  In the case of using shortened RS codes, the size of the cancelable template is considered equal to $N$ symbols rather than $N'$ symbols. For example, the output of the cancelable template block could be 768 bits which equals to $N=768/8=96$ symbols. The security of the secure multimodal template depends on the selected value of K, implying that the security of the system is k bits, where $k=mK$. The output of the decoder is a length-$k$ binary message, which is cryptographically hashed and stored as the secure multimodal template in the database. When a query is presented for authentication, the system approves the authentication only if the cryptographic hashes of the query matches with the specific enrolled identity.

\vspace{-0.30cm}
\section{Experimental Results for the cancelable multimodal template}\label{sec:CTM_EXPR}

We have evaluated the matching performance and the security of our proposed secure multibiometric system using the WVU multimodal database \cite{wvu_multimodal_2017} containing images for face and iris modalities. \comment{Note that all the experiments have been performed with optimized hyper-parameters. We have used \{$\alpha$, $\beta$, $\gamma$\} as \{8, 2, 2\} for FCA and \{6, 4, 2\} for BLA, respectively.}

In this section, we analyze the cancelable multimodal template, which is the output of the CTM. Analyzing the output of the CTM helps us to gain insight into the requirements and the strength of the error correcting code to be used in the secure sketch template module (SSTM). In the next section, we analyze the secure multimodal template, which is the output of the overall secure multimodal  system.
\vspace{-0.35cm}
\subsection{Evaluation Protocol}\label{subsec:CTMEVPR}

    For the cancelable multimodal template, equal error rate (EER) has been used as one of the metrics to evaluate the matching performance for various levels of random bit selection (values of $G$). EER indicates a value that the proportion of false acceptances is equal to the proportion of false rejections. The lower the equal error rate value, the higher the accuracy of the biometric system. We have also used the genuine and impostor distribution curves along with the receiver operating characteristic (ROC) curves to evaluate the matching performance of the cancelable template.
    
\vspace{-0.35cm}
\subsection{Performance Evaluation}\label{subsec:CTMPEEV}
After fine-tuning the entire DFB, we test this network by extracting features using the JRL of the DFB. In both the FCA and BLA architectures, the output is a 1024-dimensional joint binarized feature vector. For testing, we have used 50 subjects from the WVU-Multimodal 2012 dataset. The training and testing set are completely disjoint which means these 50 subjects have never been used in the training set. 20 face and 20 iris images are chosen randomly for each of these 50 subjects. This will give 20 pairs (face and iris) per subject with no repetitions. These 1,000 pairs $(50 \times 20 )$ are forward passed through the DFB and 1024-dimensional 1,000 fused feature vectors are extracted. A user-specific random-bit selection is performed using the fused feature vector to generate the cancelable multimodal template. The number of randomly selected bits $G$ that we have used in our experiments is equal to 128, 256, 512, 768 bits out of the 1024 dimensional binary fused vector to generate the cancelable multimodal template.

In this section, we present the results for the statistical analysis of the cancelable multimodal template, using two different architectures (FCA and BLA) for fusing the face and iris features. The performance evaluation for each architecture is also discussed here.


Two  scenarios have been considered for the evaluation of the secure templates. One is the unknown key scenario. In this scenario, the impostor does not have access to the key of the legitimate user. The impostor tries to break into the system by posing as a genuine user by presenting an artificially synthesized key (which is different from the actual key of the genuine user) and also presenting impostor biometrics. This means that the impostor will try to present random indices for our random-bit selection method in the CTM. These random indices are different from the actual indices that were selected during the enrolment for the legitimate user. The other scenario is the stolen key scenario. In this scenario the impostor has access to the actual key of the genuine user and tries to break the system by presenting actual key with impostor biometrics.
    
      \begin{figure}[t]
\centering     
\subfigure[256 bits]{\label{fig:a}\includegraphics[width=5.4cm]{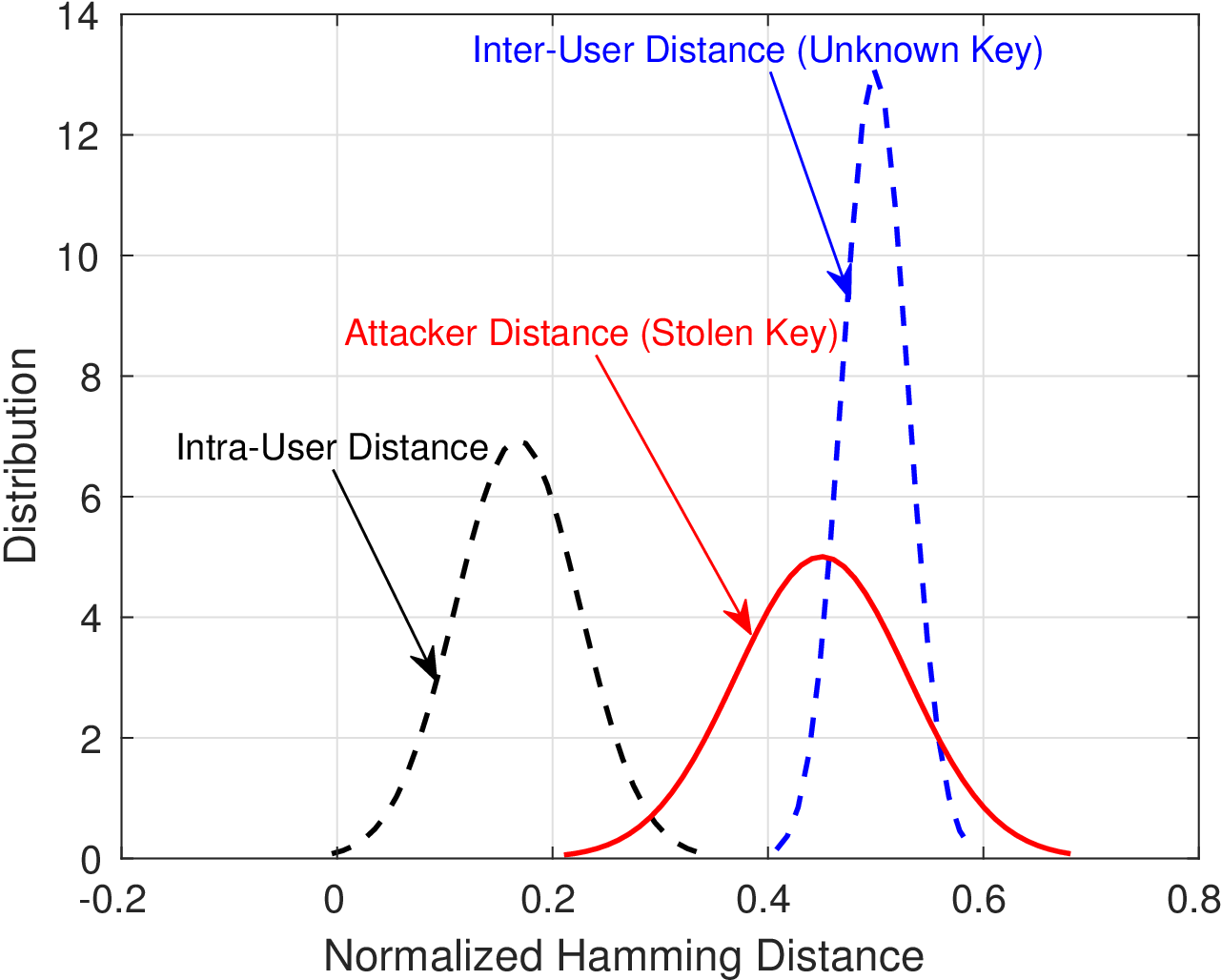}}
\subfigure[768 bits]{\label{fig:b}\includegraphics[width=5.4cm]{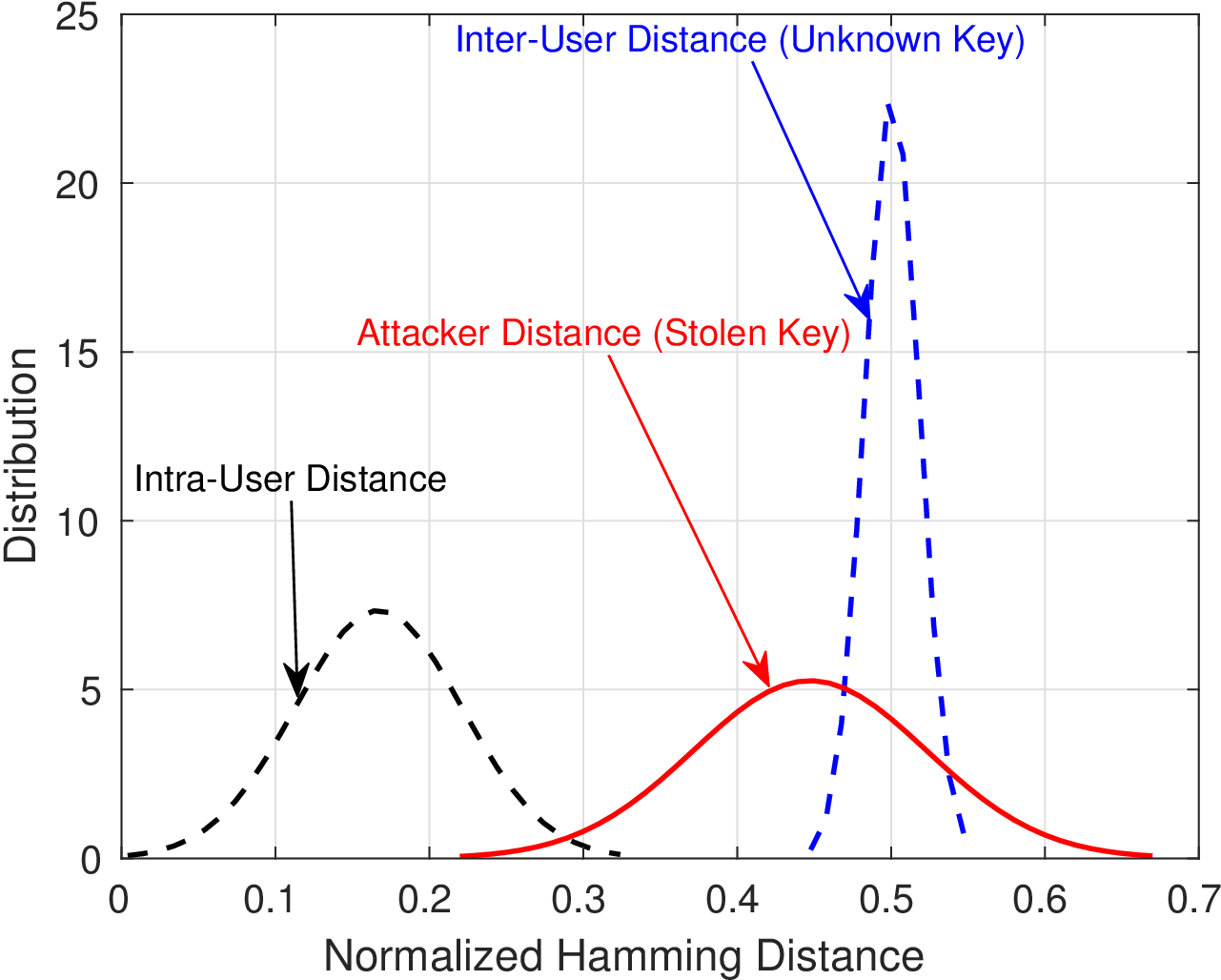}}
\subfigure[1024 bits]{\label{fig:c}\includegraphics[width=5.4cm]{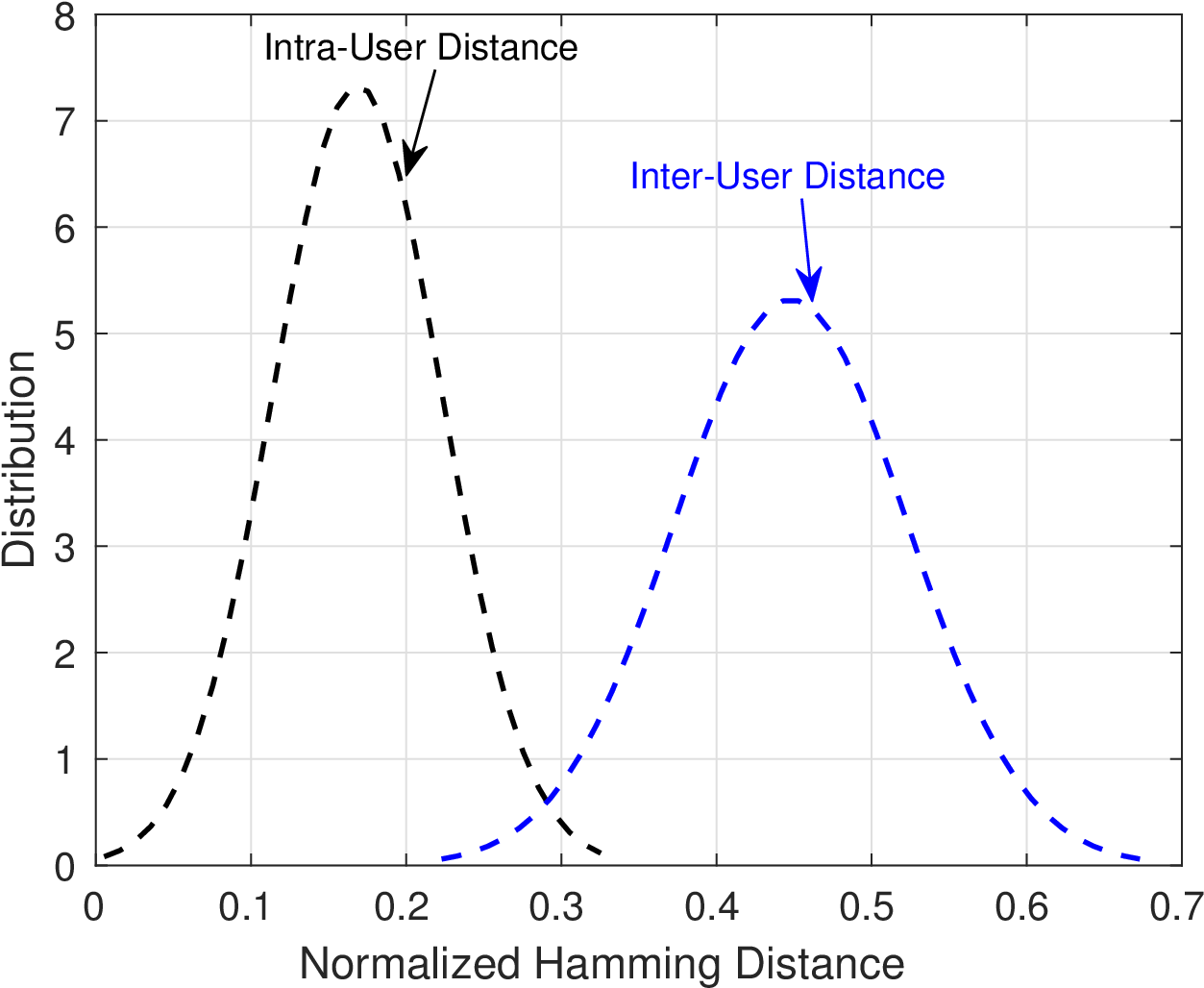}}
\caption{Genuine and impostor distribution of cancelable template distances using FCA for varying number of random bits.}
\label{fig:fca_distr_joint_unknown}
\vspace{-0.45cm}
\end{figure}


    The genuine and impostor distributions for the cancelable template for FCA in the unknown key and stolen key scenarios generated by varying the number of random bits selected by the CTM is given in Fig. \ref{fig:fca_distr_joint_unknown}. The genuine and impostor distributions shown in Fig. \ref{fig:fca_distr_joint_unknown} have been generated by fitting a normal distribution curve to the histogram. We first observe that there is no overlap between the inter-user (impostor) and intra-user (genuine) distributions. These distributions assume that every user employs his own key. Also plotted is an attacker (stolen key) distribution in which a user (attacker) uses the key of another user (victim). In this case, the attacker distribution slightly overlaps with the genuine distribution, but the overlap between the two is still reasonably small. In addition, observe that as the number of random bits selected grows from 256 to 768, the overlap between the genuine and impostor distributions reduces in both the scenarios. However, when all the 1024 bits are used, the overlap again is increased. This clearly shows the trade-off between the security (selection of 'G' random bits) and the matching performance (overlap of the distributions). Notice that there is no ``stolen key" curve in Fig. \ref{fig:c} as all the 1024 bits are used with no down-selection of bits, and hence, no key. 
    
    
    \begin{figure}[t]

\centering
\includegraphics[width=7.2cm]{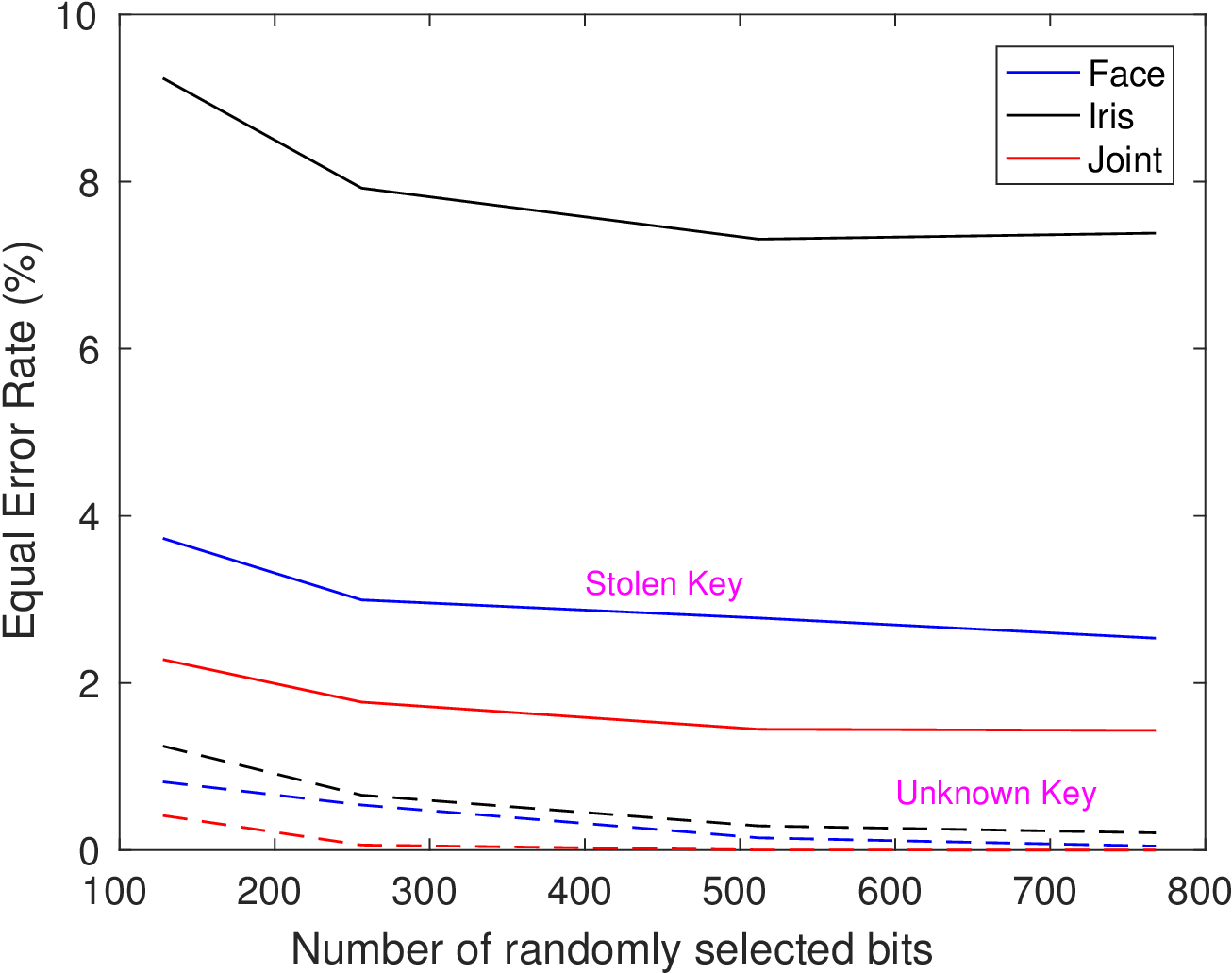}
\caption{EER curves for face, iris, joint-FCA modalities in unknown key (dashed lines) and stolen key (solid lines) scenarios using different sizes of cancelable template.}\label{fig:eer_fca_canc} 
\vspace{-0.35cm}
\end{figure}

\begin{figure}[t]

\centering
\includegraphics[width=7.2cm]{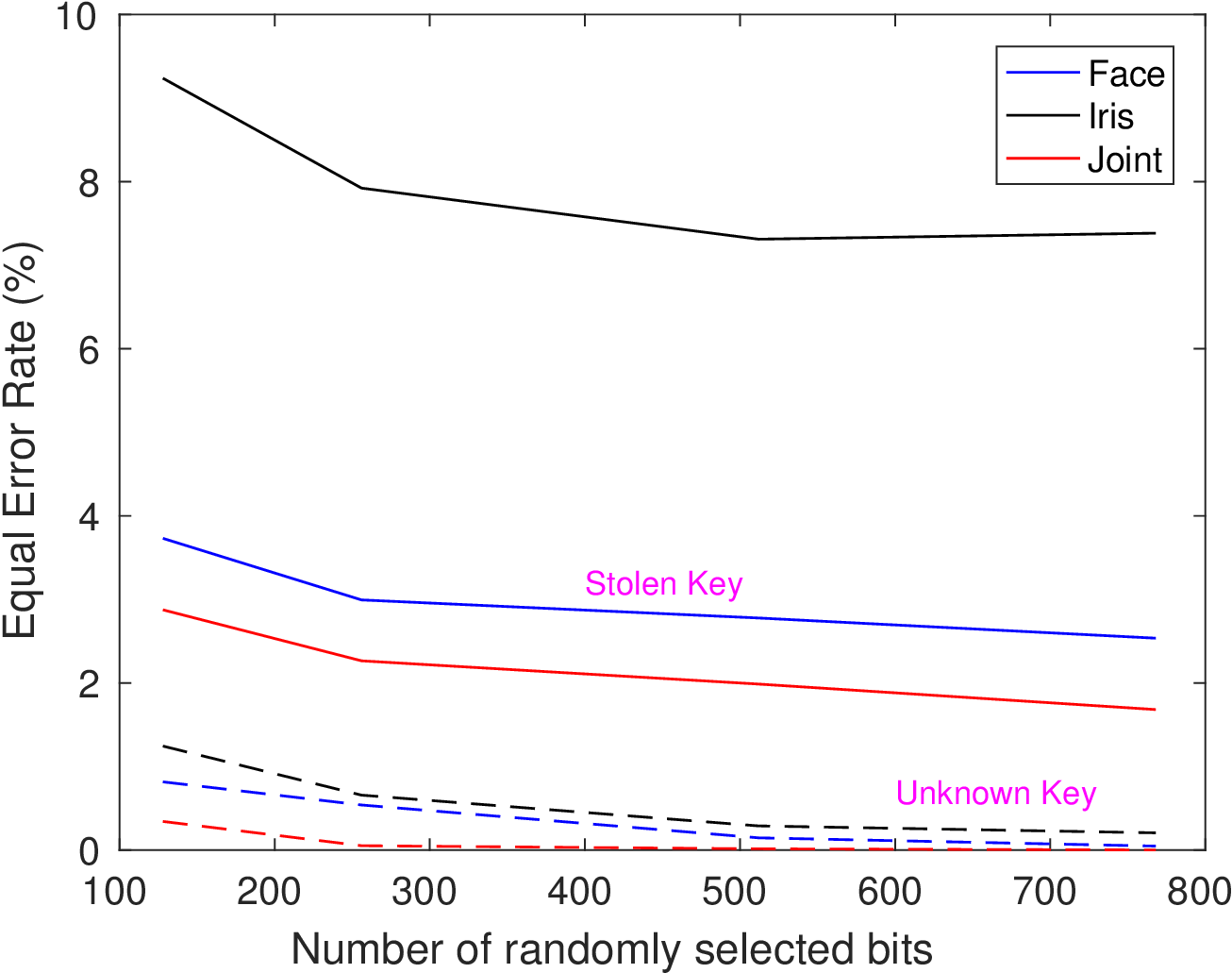}
\caption{EER curves for face, iris, and joint-BLA modalities in unknown key (dashed lines) and stolen key (solid lines) scenarios for different sizes of cancelable template.}\label{fig:eer_bla_canc} 
\vspace{-0.45cm}
\end{figure}
    


    



The EER plots for FCA and BLA are given in Fig. \ref{fig:eer_fca_canc} and Fig. \ref{fig:eer_bla_canc}, respectively. EER plot is obtained by calculating the value of EER by varying the length of the cancelable template (number of randomly selected bits). In general, it can be observed from the EER plots that there is an increase in performance by using additional biometric features and the multimodality (joint) template performs better than the individual modalities (face and iris). As seen from the curves, the EER for the joint modality is lower than the EER for face or iris. For example, the EER for joint modality using FCA and BLA at 512 bits for stolen key scenario is $1.45\%$ and $1.99\%$, respectively. Using the same settings, the EER for face and iris is $2.6\%$ and $7.4\%$, respectively. This clearly shows that there is an improvement by fusing multiple modalities.

\begin{figure}

\centering
\includegraphics[width=7.3cm]{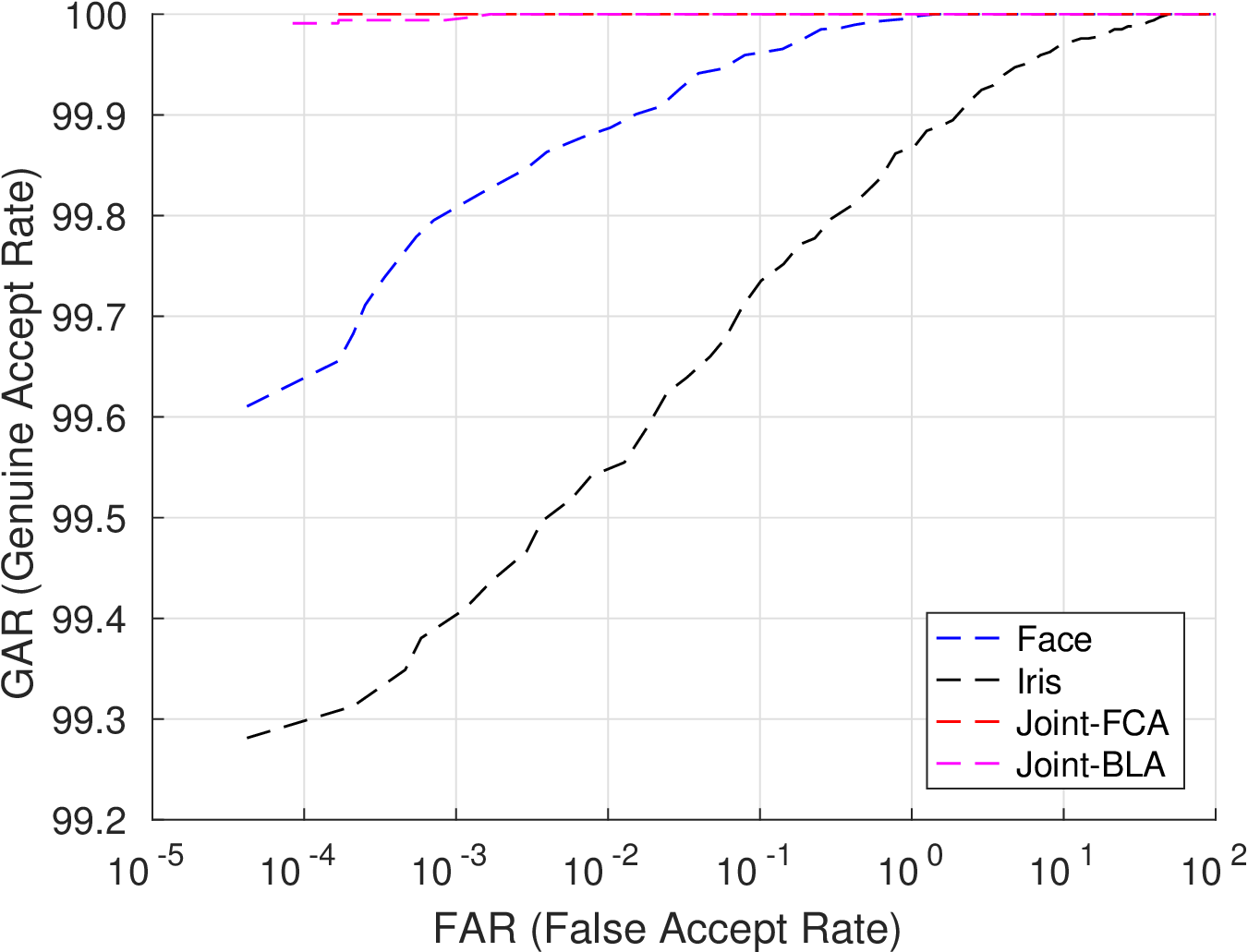}
\caption{ROC curves for face, iris, joint-FCA, and joint-BLA in unknown key scenario for a random selection of 768 bits.}\label{fig:ROC_fca_bla_canc_unknown} 
\vspace{-0.45cm}
\end{figure}

\begin{figure}

\centering
\includegraphics[width=7.3cm]{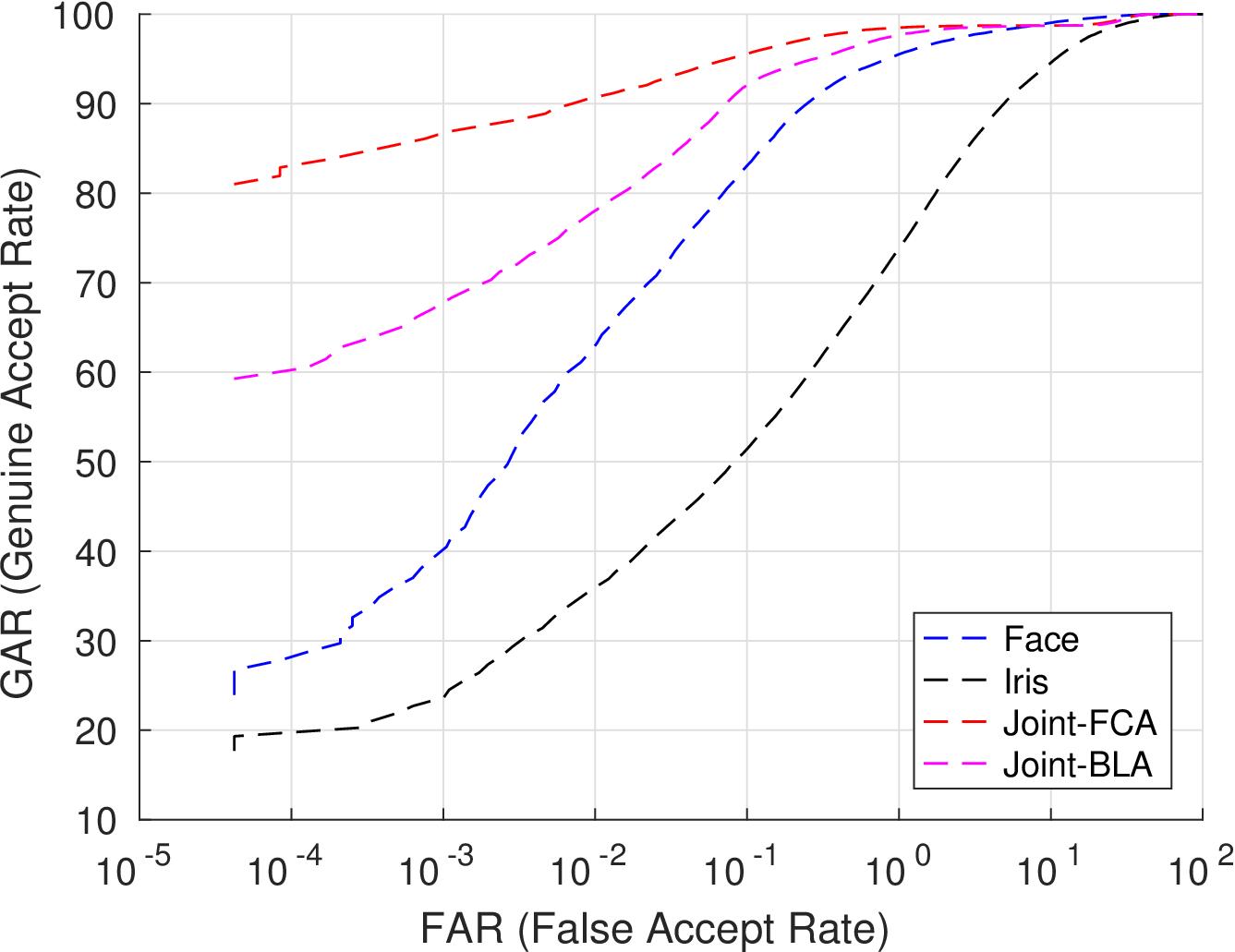}
\caption{ROC curves for face, iris, joint-FCA, and joint-BLA in stolen key scenario for a random selection of 768 bits.}\label{fig:ROC_fca_bla_canc_stolen} 
\vspace{-0.45cm}
\end{figure}

The  ROC curves for both the architectures have been compared in Fig. \ref{fig:ROC_fca_bla_canc_unknown} and \ref{fig:ROC_fca_bla_canc_stolen} for unknown and stolen key scenarios, respectively, when the number of randomly selected values (security) is 768 bits. Again, we can clearly observe that the joint modality performs better than the individual modality. For a false accept rate (FAR) of $0.5\%$, the genuine accept rate (GAR) for stolen key scenario using FCA and BLA is $98.25\%$ and $96.33\%$, respectively. For face and iris, the GAR is $90.8\%$ and $62.5\%$, respectively at an FAR of $0.5\%$. 

As observed from the plots, the matching performance is not compromised for high security and the multimodality gives us better performance than unimodality.

\section{Experimental Results for the Overall System}\label{sec:SSTM_EXPR}

In this section, we analyze the performance at the output of the overall system, where the output of the overall system is the secure multimodal template that is stored in the database. 
\vspace{-0.75cm}
\subsection{Evaluation Protocol}\label{subsec:SSTMEVPR}
 We evaluate the trade-off between the matching performance and the security of the proposed secure multimodal system using the curves that relate the GAR to the security in bits (i.e., the G-S curves). The G-S curve is acquired by varying the error correcting capability of the Reed-Solomon code used for FEC decoding in the SSTM. The error correcting capability of a code signifies the number of bits (or symbols) that a given ECC can correct. The error correcting capability of a Reed-Solomon code is given by $\frac{(N-K)}{2}$ symbols or $\frac{(n-k)}{2}$ bits. We vary the error correcting capability of the code by using different code rates ($K/N$).
    
\vspace{-0.35cm}
\subsection{Performance Evaluation}\label{subsec:SSTMPEEV}
As explained in Sec. \ref{subsec:exptsetup}, the output of the cancelable template block ($n$ bits) is decoded in order to generate a multimodal secure sketch of length $k$ bits, where $k$ also represents the security of the proposed secure multibiometric system. This multimodal sketch is cryptographically hashed and stored as the secure multimodal template in the database. When a query is presented for authentication, the system authenticates the user only if the cryptographic hash of the query matches that of the specific enrolled identity.

\begin{figure}

\centering
\includegraphics[width=7.2cm]{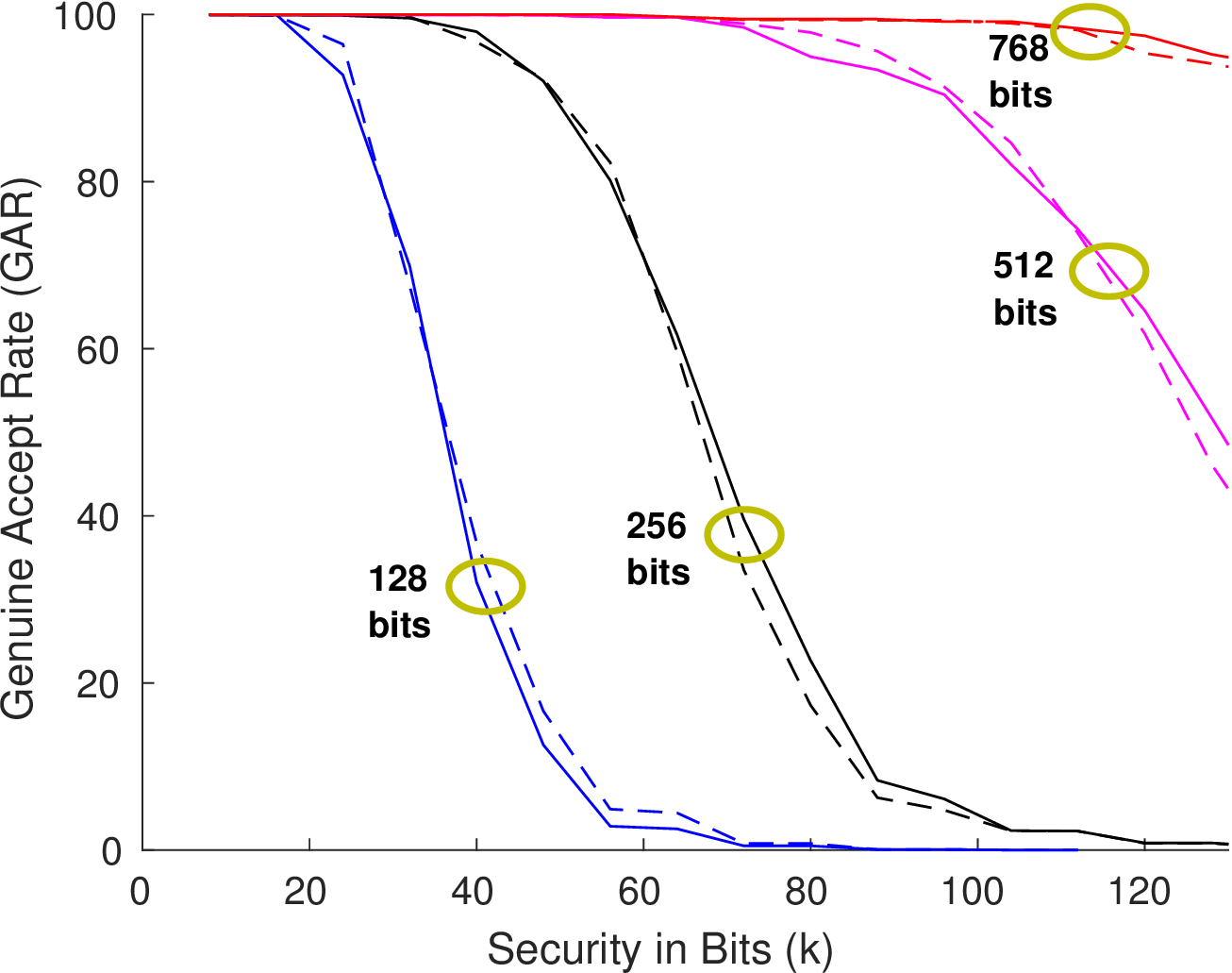}
\caption{G-S curves using FCA in unknown key (dashed) and stolen key (solid) scenarios for different values of $n$ bits. }\label{fig:GS_fca_joint_both} 
\vspace{-0.35cm}
\end{figure}

\begin{figure}

\centering
\includegraphics[width=6.8cm]{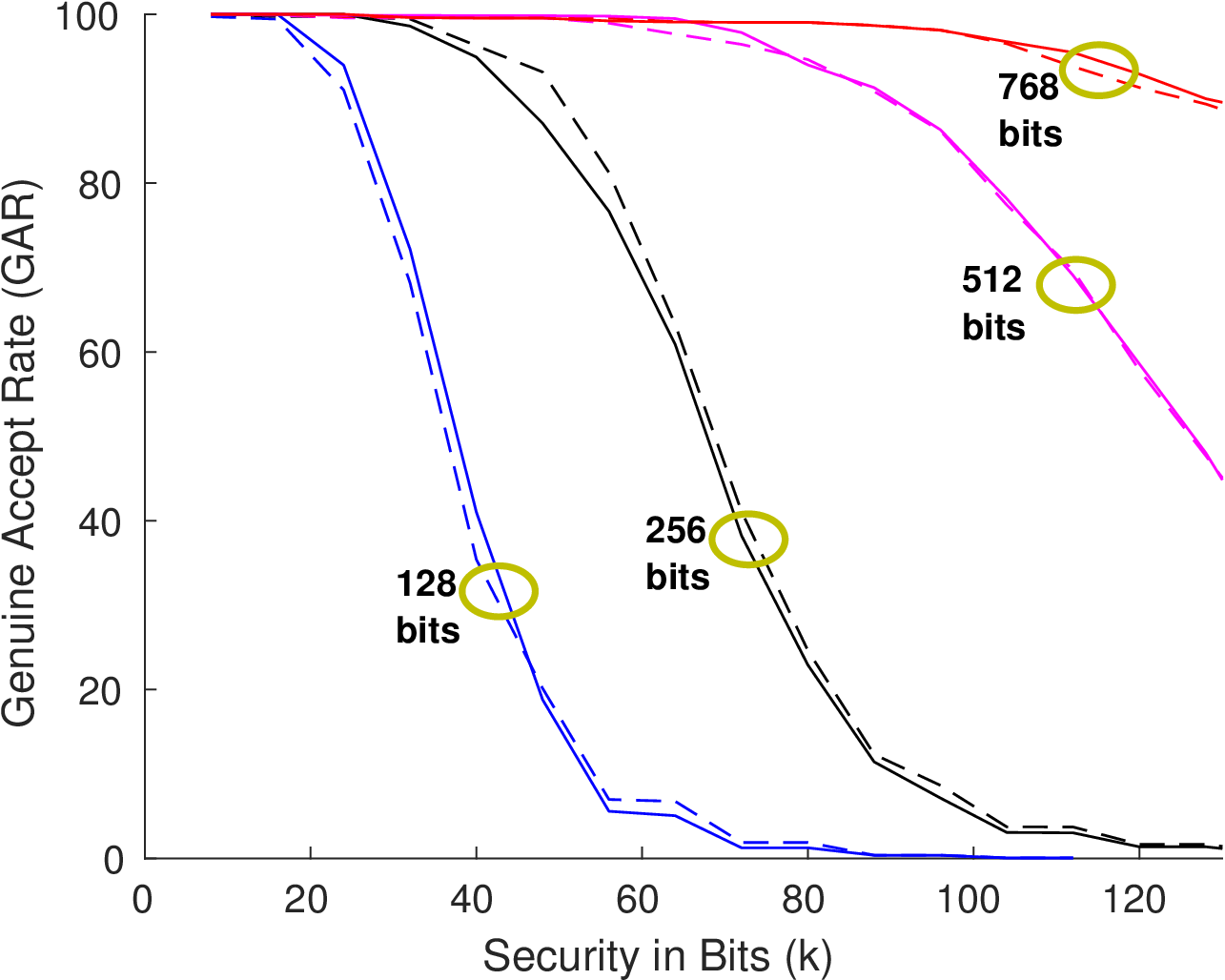}
\caption{G-S curves using BLA in unknown key (dashed) and stolen key (solid) scenarios for different values of $n$ bits. }\label{fig:GS_bla_joint_both} 
\vspace{-0.35cm}
\end{figure}

We have experimented with different values of $N$ symbols with $m=8$ and $N'=255$ symbols using shortened RS codes. The G-S curves for different values of $n$ bits (equivalent to $N$ symbols) for unknown and stolen key scenarios using FCA and BLA are given in Fig. \ref{fig:GS_fca_joint_both} and Fig. \ref{fig:GS_bla_joint_both}, respectively. We can observe from the curves that as the size of the cancelable template in bits ($n$) increases, the GAR for a given level of security in bits ($k$) also increases. 

\begin{table*}[h]
\centering
\captionsetup{width=.75\linewidth}
\caption{GARs of FCA and BLA in unknown and stolen key scenarios at a security level of 56, 80 and 104 bits using different cancelable template size ($N$).}
\scalebox{0.70}{\begin{tabular}{|c|p{0.40cm}|c|c|c|c|c|c|c|}
 \hline
\multicolumn{1}{|c}{\multirow{2}{*}{$N$}} &\multicolumn{1}{|c}{\multirow{2}{*}{$n$}} &\multicolumn{1}{|c}{\multirow{2}{*}{Security ($K$)}}&\multicolumn{1}{|c}{\multirow{2}{*}{Security ($k$) }} &\multicolumn{1}{|c}{\multirow{2}{*}{$\frac{(n-k)}{2}$}} &\multicolumn{2}{|c|}{FCA-GAR} &\multicolumn{2}{|c|}{BLA-GAR}\\ [0.5ex] 

 \cline{6-9}
 (symbols)&(bits) &(symbols)&  (bits)  &  & Unknown & Stolen & Unknown & Stolen \\ \hline \hline
 \multirow{3}{*}{32} & \multirow{3}{*}{256} &7& 56 & 100 & 82.30\% & 82.15\% & 82.25\% & 80.66\% \\ \cline{3-9}
& &10& 80 & 88 & 31.32\% & 32.68\% & 36.67\% & 35.92\% \\ \cline{3-9} 
& &13& 104 & 76 & 4.3\% & 4.33\%  & 6.77\% & 6.07\% \\ \hline \hline
\multirow{3}{*}{64} & \multirow{3}{*}{512} &7& 56 & 228 & 99.65\% & 99.68\% & 98.95\% & 99.77\% \\ \cline{3-9}
& &10& 80 & 216 & 97.85\% & 94.95\% & 94.63\% & 94\% \\ \cline{3-9} 
& &13& 104 & 204 & 84.63\% & 82.05\%  & 84.41\% & 85.15\% \\ \hline \hline
\multirow{3}{*}{96} & \multirow{3}{*}{768} &7& 56 & 356 & 99.93\% & 99.99\% & 99.55\% & 99.22\% \\ \cline{3-9}
& &10& 80 & 344 & 99.37\% & 99.44\% & 99.04\% & 99.04\% \\ \cline{3-9} 
& & 13&104 & 332 & 98.95\% & 99.16\% & 96.51\% & 96.75\% \\ \hline 


\end{tabular}}

\label{table:GAR_FCA_BLA}
\end{table*}

For example at a security ($k$) of $104$ bits (equivalent to $K=13$ symbols) using FCA with the stolen key scenario, the GAR for $n$=128, 256, 512, and 768 bits is equal to $0.62\%,4.33\%,82.05\%$, and $99.16\%$, respectively. Similarly for the unknown key scenario and FCA, the GAR for $n$=128, 256, 512, and 768 bits is equal to $0.74\%,4.3\%,84.63\%$, and $98.95\%$, respectively. It can be observed that the use of a larger cancelable template results in better performance. This performance improvement can be attributed to the fact that an increase in $n$ at a fixed value of $k$ (security) improves the error correcting capability of the RS codes which is given by $\frac{(n-k)}{2}$ and hence a better matching performance. 


Table \ref{table:GAR_FCA_BLA} summarizes the  GAR for different values of $n$ at security levels of 56, 80, and 104 bits using both FCA and BLA. The error correcting capabilities in bits $\left(\frac{(n-k)}{2}\right)$ for the RS codes at different security levels are also given in the table. From the Table \ref{table:GAR_FCA_BLA},  it can be observed that for a given size of the cancelable template in bits ($n$), the error correcting capability decreases with an increase in the required security levels in bits ($k$) of the system, which results in a decrease in GAR. This implies that the code cannot  correct the intra-class variations at high code rates ($k/n$) (higher value of $k$), which results in a reduced GAR. This is the trade-off between the matching performance (GAR) and the security ($k$) of the system. We have chosen a minimum security level of 56 bits for comaprison in Table \ref{table:GAR_FCA_BLA} which is higher when compared to those reported in the literature \cite{nagar_multibiometriccryptosystems_2012}.







\begin{figure}

\centering
\includegraphics[width=6.8cm]{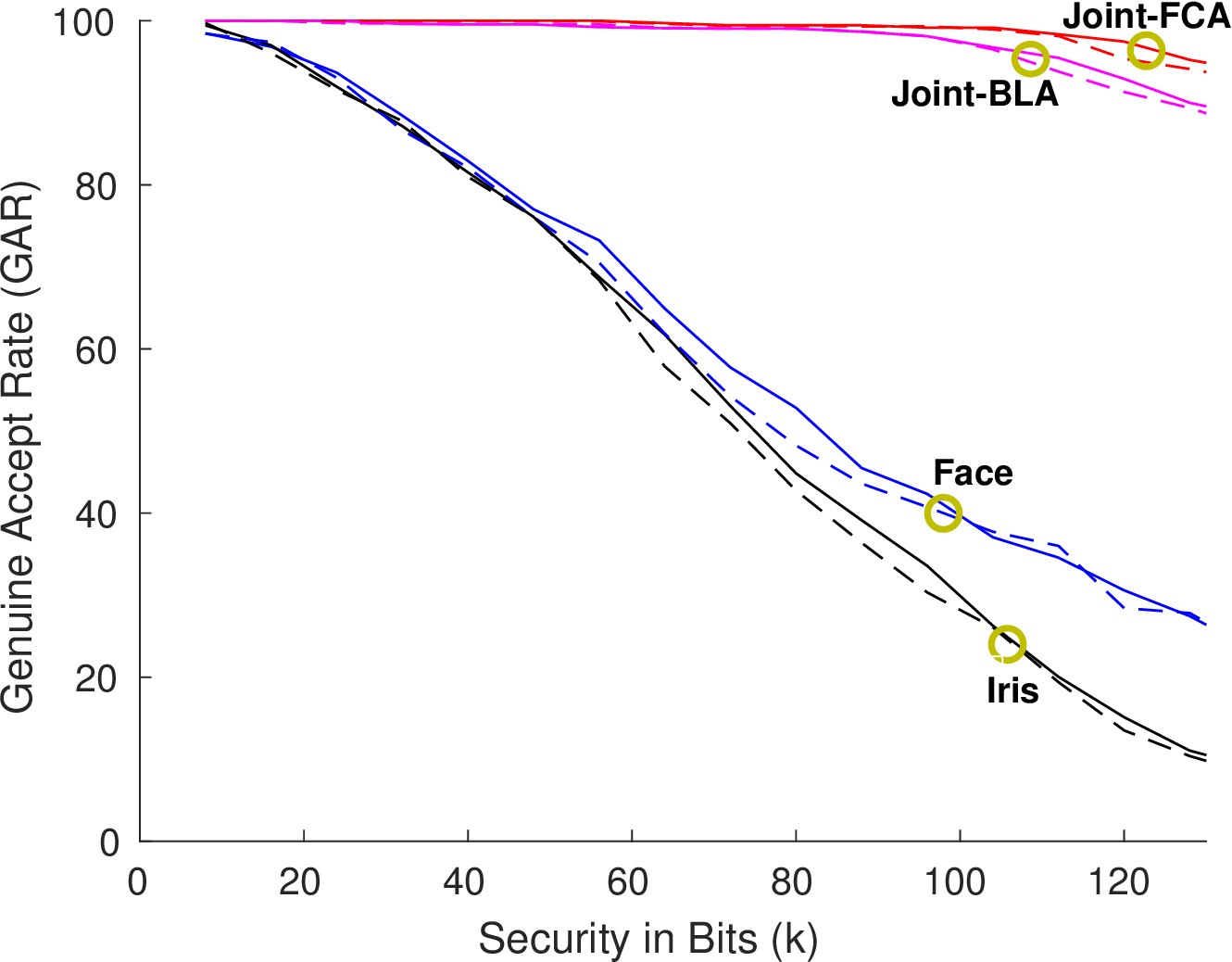}
\caption{G-S curves for face, iris, joint-FCA, and joint-BLA  modalities in unknown key (dashed lines) and stolen key (solid lines) scenarios for $n=768$ bits.}\label{fig:GS_comp_unknown_stolen} 
\vspace{-0.35cm}
\end{figure}

The plot in Fig. \ref{fig:GS_comp_unknown_stolen} gives a comparison of G-S curves for face, iris, joint-FCA, and joint-BLA modalities using $m=8, N'=255$ and $n=768$ bits (equivalent to $N=96$ symbols) for unknown and stolen key scenario, respectively. The security for the iris modality in stolen key scenario at a GAR of $95\%$ is 20 bits. However, by incorporating additional biometric features (face), the security of the multibiometric system using FCA increases to 128 bits at the same GAR.


\vspace{-0.45cm}

\comment{\subsection{Comparison with State-of-the-Art Hashing Techniques}}

\begin{figure}[t]
\centering
\includegraphics[width=6.6cm]{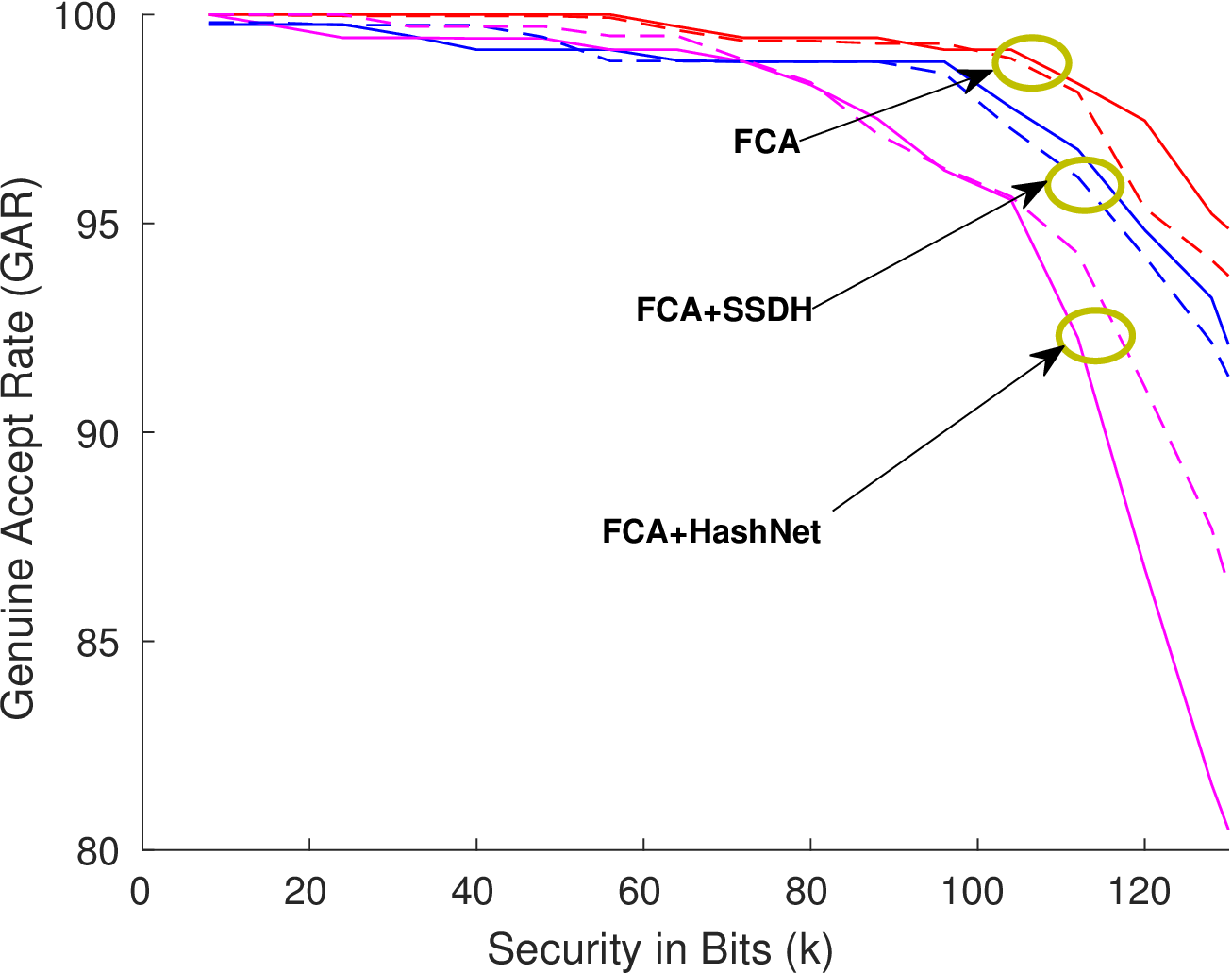}
\caption{\comment{G-S curves to compare performance of the proposed hashing with two other hashing techniques for FCA  in unknown (dashed lines) and stolen key (solid lines) scenarios.}}\label{fig:GS_SSDH_Hashnet_both} 
\vspace{-0.50cm}
\end{figure}

\comment{As a further experiment, we compare the proposed hashing technique with other hashing techniques. This is done by replacing our hashing method with two other hashing methods \cite{ssdh_2018_pami} and \cite{cao_2017_hashnet}, and then training and testing the multimodal authentication system using the same WVU multimodal dataset. The rest of the system is kept the same for comparison purposes. We have compared our hashing technique with supervised semantics-preserving deep hashing (SSDH) \cite{ssdh_2018_pami}, and HashNet \cite{cao_2017_hashnet} and evaluated the overall system to produce G-S curves. We have used the FCA system for comparison. We denote the system with our proposed hashing technique as ``FCA", use ``FCA+SSDH" to denote the FCA architecture with our hashing function replaced by the SSDH hashing, and use ``FCA+HashNet" to denote our FCA architecture with the HashNet hashing function. Fig. \ref{fig:GS_SSDH_Hashnet_both} shows G-S curves for stolen key and unknown key scenarios. It can clearly be seen that our proposed hashing method performs better than the other two deep hashing techniques for the given multimodal biometric security application. Compared to the other two hashing techniques, our proposed method improves the GAR by at least $1.15\%$ at a high security of 104 bits. A comparison of our hashing technique against others for an image-retrieval application can be found in the Appendix.} 


\vspace{-0.35cm}

\comment{\subsection{Ablation Study}\label{subsec:hyper_tun}
The objective function defined in (\ref{eq:6}) contains 3 constraints, one for the semantics-preserving binary codes (i.e., for classification) and two constraints for efficient binary codes (i.e., for binarization and entropy maximization). In this section, we study the relative importance of each of these terms.} 

\comment{First, we measure the influence of the classification term $E_1$ by setting $\alpha = 1$, $\beta = 0$, and $\gamma = 0$. Using this setting, we train our DFB model and evaluate the overall system by calculating the GAR for a security of $k=$ 56, 80, and 104 bits for $n=768$ bits (similar to Table \ref{table:GAR_FCA_BLA}) on the test data for the WVU-Multimodal 2012 dataset. We also study the effect of the binarization constraint along with classification term by setting $\alpha = 1$, $\beta = 1$, and $\gamma = 0$, train our DFB model and again evaluate the overall system by calculating the GARs. Finally, we set $\alpha = 1$, $\beta = 1$, and $\gamma = 1$, and train the DFB model and evaluate the overall system. We performed this experiment for both FCA and BLA architectures only for stolen key scenario because we can see from Table \ref{table:GAR_FCA_BLA} that unknown key and stolen key scenarios give very similar results. The GAR results for this experiment are shown in Table \ref{table:GAR_hyper}}. 

\comment{It can be observed from Table \ref{table:GAR_hyper} that the classification term $E_1$ is the most important term. However, adding the binarization and the entropy constraints $E_2$ and $E_3$ (i.e., $\alpha=1, \beta=1, \gamma=1$) definitely help to improve the matching performance (i.e., GAR) by at least $1.25\%$ at a high security of 104 bits in our proposed system. We also note that this performance improvement is evident for both FCA and BLA architectures. Therefore, using all the terms proves beneficial to improve the matching performance evident at higher level of security for both FCA and BLA architectures.}

\begin{table}[h]
\centering
\captionsetup{width=\linewidth}
\caption{\comment{GARs of FCA and BLA in stolen key scenario showing the influence of each term in the objective function.}}
\scalebox{0.72}{\begin{tabular}{|c|p{0.50cm}|c|c|c|}
 \hline
\multicolumn{1}{|c}{} &\multicolumn{1}{|c}{\multirow{2}{*}{$n$}} &\multicolumn{1}{|c}{\multirow{2}{*}{Security ($k$) }} &\multicolumn{1}{|c}{} &\multicolumn{1}{|c|}{}\\ [0.5ex] 
 Hyper-parameters&(bits)&  (bits)  & FCA-GAR & BLA-GAR  \\ \hline \hline
\multirow{3}{*}{$\alpha=1,\beta=0,\gamma=0$} & \multirow{3}{*}{768}& 56& 99.16\% & 98.71\% \\ \cline{3-5}
& & 80 & 98.32\%  & 96.87\% \\ \cline{3-5} 
& & 104 & 95.26\% & 93.29\% \\ \hline\hline 
\multirow{3}{*}{$\alpha=1,\beta=1,\gamma=0$} & \multirow{3}{*}{768}& 56& 99.73\% & 98.76\% \\ \cline{3-5}
& & 80 & 98.8\%  & 97.14\% \\ \cline{3-5} 
& & 104 & 95.72\% & 94.72\% \\ \hline \hline
\multirow{3}{*}{$\alpha=1,\beta=0,\gamma=1$} & \multirow{3}{*}{768}& 56& 99.52\% & 98.70\% \\ \cline{3-5}
& & 80 & 98.41\%  & 97.02\% \\ \cline{3-5} 
& & 104 & 95.43\% & 93.98\% \\ \hline \hline
\multirow{3}{*}{$\alpha=1,\beta=1,\gamma=1$} & \multirow{3}{*}{768}& 56& 99.9\% & 99\% \\ \cline{3-5}
& & 80 & 99.8\%  & 97.6\% \\ \cline{3-5} 
& & 104 & 96.5\% & 95.6\% \\ \hline
\end{tabular}}
\label{table:GAR_hyper}
\vspace{-0.25cm}
\end{table}

\vspace{-0.40cm}

\subsection{Privacy Analysis}

The objective of our work is to design a multimodal authentication system that maximizes the matching performance while keeping the biometric data secure. However, the problem is complicated by the possibility that the adversary may gain access to the enrollment key $\textbf{k}_\textbf{e}$, the multimodal secure sketch $\textbf{s}_\textbf{e}$, the enrollment feature vector $\textbf{e}$, or any combination thereof. Using this information, the adversary could not only compromise the authentication integrity of the system, but may also extract information about the biometric data. The system should be robust in these scenarios and the system design should minimize the privacy leakage, which is the leakage of the user biometric information from the compromised data, and preserve authentication integrity of the system.

The G-S curves which have been discussed in Sec. \ref{subsec:SSTMPEEV} quantify the security of the system. In this subsection, we will quantify the privacy leakage of the user's biometric information for our proposed system. The privacy of the user is compromised if the adversary gains access to the enrollment feature vector \textbf{e} as we assume that the enrollment feature vector can be de-convolved to recover the biometric data of the user. The information leaked about the user's enrollment feature vector $\textbf{e}$ can be quantified as mutual information:
\vspace{-0.20cm}
\begin{equation} I(\textbf{e};\textbf{V})=H(\textbf{e})-H(\textbf{e}|\textbf{V}), \label{eq:7}\end{equation} where $\textbf{e}$ represents the enrollment feature vector, and $\textbf{V}$ represents the information that adversary has access to. \textbf{V} could be the enrollment key $\textbf{k}_\textbf{e}$ and/or the multimodal secure sketch $\textbf{s}_\textbf{e}$. $H(\textbf{e})$ represents entropy of \textbf{e} and quantifies the number of bits required to specify $\textbf{e}$. In particular, $H(\textbf{e})=J$ because the optimization described in Sec. \ref{subsec:obj} is designed to ensure that the $J$ bits in the encoded template are independent and equally likely to be 0 or 1. $H(\textbf{e}|\textbf{V})$ is the entropy of \textbf{e} given \textbf{V} and quantifies the remaining uncertainty about $\textbf{e}$ given knowledge of $\textbf{V}$. The mutual information $I(\textbf{e};\textbf{V})$ is the reduction in uncertainty about $\textbf{e}$ given $\textbf{V}$ \cite{rane_secure_biom_2013}. 

 Let's assume that the adversary gains access to the enrollment key $\textbf{k}_\textbf{e}$. In this case $\textbf{V}=\textbf{k}_\textbf{e}$ and mutual information is: 
\vspace{-0.20cm}
\begin{equation} I(\textbf{e}; \textbf{k}_\textbf{e})=H(\textbf{e})-H(\textbf{e}| \textbf{k}_\textbf{e})=0, \label{eq:8}\end{equation} because $H(\textbf{e}| \textbf{k}_\textbf{e})=H(\textbf{e})=J$ as the key $\textbf{k}_\textbf{e}$ does not give any information about the enrollment feature vector \textbf{e}. $\textbf{k}_\textbf{e}$ just gives the indices of the random values selected from \textbf{e} but does not provide values at those indices. 


The information leakage when $\textbf{s}_\textbf{e}$ or the pair $(\textbf{k}_\textbf{e},\textbf{s}_\textbf{e})$ is compromised can be quantified using the conditional mutual information because $\textbf{s}_\textbf{e}$ is dependent on $\textbf{r}_\textbf{e}$ which is driven by $\textbf{k}_\textbf{e}$. Hence, the information leakage when the secure sketch is compromised is conditionally dependent on $\textbf{k}_\textbf{e}$ and given as:
\vspace{-0.20cm}
\begin{equation} I(\textbf{e}; \textbf{s}_\textbf{e}|\textbf{k}_\textbf{e})=H(\textbf{e}| \textbf{k}_\textbf{e})-H(\textbf{e}| \textbf{s}_\textbf{e},\textbf{k}_\textbf{e}), \label{eq:9}\end{equation} where $H(\textbf{e}| \textbf{k}_\textbf{e})$ quantifies the remaining uncertainty about $\textbf{e}$ given knowledge of $\textbf{k}_\textbf{e}$ and $H(\textbf{e}| \textbf{s}_\textbf{e},\textbf{k}_\textbf{e})$ quantifies the remaining uncertainty about $\textbf{e}$ given knowledge of $\textbf{k}_\textbf{e}$ and $\textbf{s}_\textbf{e}$. This conditional mutual information is measured under two scenarios discussed below.

 \textbf{Both $\textbf{s}_\textbf{e}$ and $\textbf{k}_\textbf{e}$ are compromised:} In this scenario the adversary gains access to both $\textbf{s}_\textbf{e}$ and $\textbf{k}_\textbf{e}$. As previously discussed, $H(\textbf{e}| \textbf{k}_\textbf{e})=H(\textbf{e})=J$ because knowing $\textbf{k}_\textbf{e}$ does not provide any information about $\textbf{e}$. If the adversary knows $\textbf{s}_\textbf{e}$, the information leakage of $\textbf{r}_\textbf{e}$ due to $\textbf{s}_\textbf{e}$ is equal to the length of $\textbf{s}_\textbf{e}$ which is $k$ bits. The adversary can use this information of $\textbf{r}_\textbf{e}$ with the additional knowledge of the enrollment key $\textbf{k}_\textbf{e}$ and exactly know the indices  and the values for the $k$ bits in the enrollment vector $\textbf{e}$. However, there is still uncertainity about the remaining $J-k$ bits of the enrollment feature vector $\textbf{e}$, which implies $H(\textbf{e}| \textbf{s}_\textbf{e},\textbf{k}_\textbf{e})=J-k$. Therefore, the information leakage about enrollment feature vector when both secure sketch and enrollment key are compromised is:
 \vspace{-0.15cm}
   \begin{equation}
\begin{split}
     I(\textbf{e}; \textbf{s}_\textbf{e}|\textbf{k}_\textbf{e})&=H(\textbf{e}| \textbf{k}_\textbf{e})-H(\textbf{e}| \textbf{s}_\textbf{e},\textbf{k}_\textbf{e})\\&=J-(J-k)=k.
    \end{split}
\label{eq:10}\end{equation}

\textbf{Only $\textbf{s}_\textbf{e}$ is compromised:} In this scenario the adversary gains access to only $\textbf{s}_\textbf{e}$. Even in this case if the adversary knows $\textbf{s}_\textbf{e}$, the information leakage of $\textbf{r}_\textbf{e}$ due to $\textbf{s}_\textbf{e}$ is $k$ bits. However, the adversary does not have any information about the enrollment key $\textbf{k}_\textbf{e}$ which means that there is added uncertainty in the information about the enrollment feature vector \textbf{e} as the adversary does not know the exact locations of the $k$ bits given by $\textbf{s}_\textbf{e}$.  This added uncertainity is measured by $H(\textbf{k}_\textbf{e})$ which is calculated using combinatorics and is:
\vspace{-0.15cm}
\begin{equation}
    H(\textbf{k}_\textbf{e})=\log_{2}\binom{J}{n},
\label{eq:11}\end{equation} where $n$ is the size of the key and (\ref{eq:11}) provides all the combinations that $n$ bits could be selected from $J$. Therefore, the conditional mutual information is given as:
\vspace{-0.10cm}
\begin{equation}
\begin{split}
     I(\textbf{e}; \textbf{s}_\textbf{e}|\textbf{k}_\textbf{e})&=H(\textbf{e}| \textbf{k}_\textbf{e})-H(\textbf{e}|
 \textbf{s}_\textbf{e},\textbf{k}_\textbf{e})\\&=J-\left(J-k+\log_{2}\binom{J}{n}\right)\\&=k-\log_{2}\binom{J}{n}\\&=\mbox{max}\left(0,k-\log_{2}\binom{J}{n}\right),
    \end{split}\label{eq:12}
\end{equation} where the $\mbox{max}$ function is applied in the last equation as information leakage cannot be negative. We have evaluated (\ref{eq:12}) using different values of $n$ and $k$ for $J=1024$ bits. We know that $n$ ranges from $1$ to $J$ depending on the number of random bits selected from the enrollment feature vector \textbf{e} and $k$ ranges from $1$ to $n$ depending on the rate of the error correcting code. We found that information leakage is zero for all the values of $k$ for $n$ ranging from $1$ to $792$ bits. However, if $n > 792$, there is a positive information leakage for $k > 780$.


From (\ref{eq:10}) and (\ref{eq:12}), we can conclude that for $J=1024$, the ideal value of $n$ should be less than $792$ and ideal value of $k$ should be small. This would make the information leakage to be zero or small in case if $\textbf{s}_\textbf{e}$ or the pair $(\textbf{s}_\textbf{e},\textbf{k}_\textbf{e})$ gets compromised. These values of $n$ and $k$ would also keep the matching performance high as shown in Fig. \ref{fig:GS_comp_unknown_stolen}. 

\vspace{-0.30cm}
\subsection{\commentt{Unlinkability Analysis}}

\commentt{According to ISO/IEC International Standard 24745 \cite{ISO24745}, transformed templates generated from the same biometric references should not be linkable across applications or databases. By using the protocol defined in \cite{unlinkability_analysis}, we have evaluated the unlinkability of the proposed system. The protocol in \cite{unlinkability_analysis} is based on mated ($H_{m}$) and non-mated ($H_{nm}$) samples distributions. Mated samples correspond to the templates extracted from the samples of the same subject using different user-specific keys. Non-mated samples correspond to the templates extracted from the samples of different subjects using different keys. For an unlinkable system,
there must exist a significant overlap between mated and non-mated score distributions \cite{unlinkability_analysis}}. 

\begin{figure}[t]
\centering      
\subfigure[FCA-104]{\label{fig:d}\includegraphics[width=4.3cm]{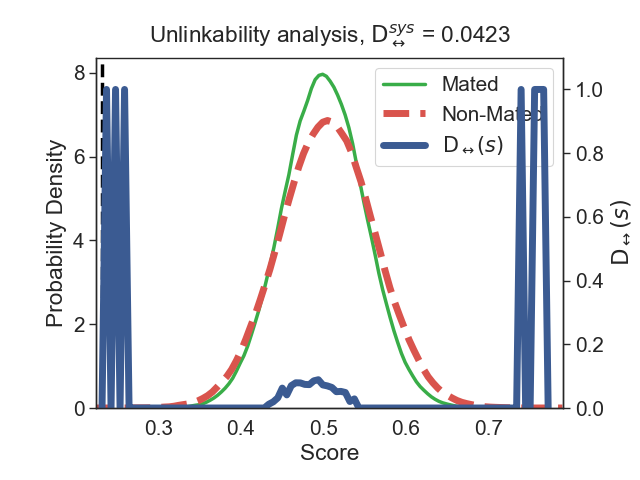}}
\subfigure[FCA-128]{\label{fig:e}\includegraphics[width=4.3cm]{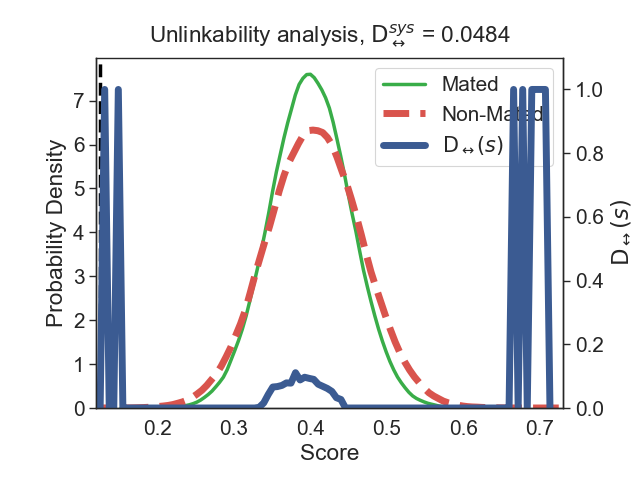}}
\subfigure[BLA-104]{\label{fig:f}\includegraphics[width=4.3cm]{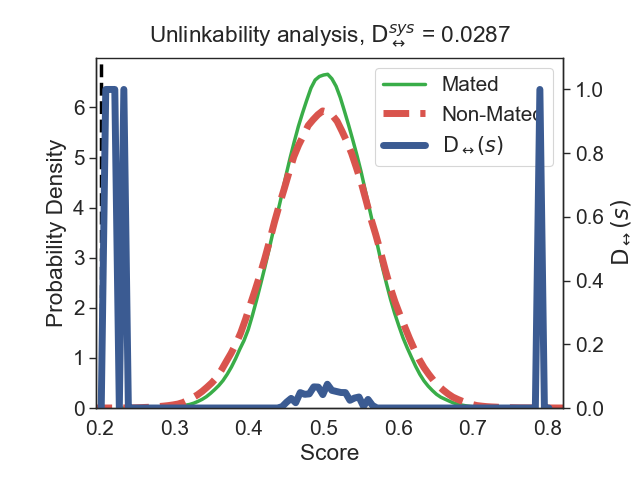}}
\subfigure[BLA-128]{\label{fig:g}\includegraphics[width=4.3cm]{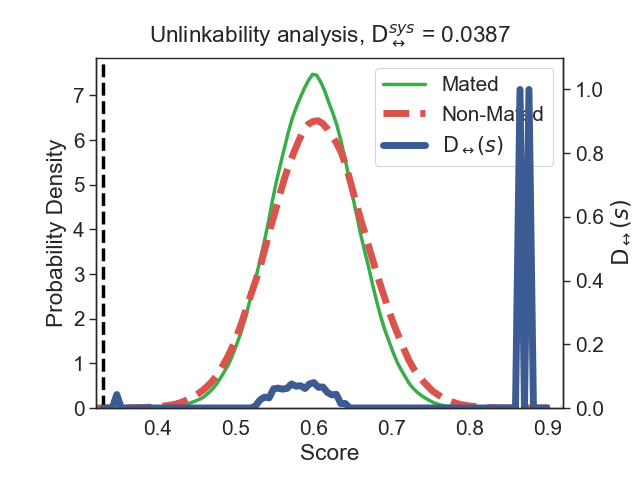}}
\caption{\commentt{Unlinkability analysis of the proposed system for FCA and BLA for different quantities of security bits (104, 128).}}
\label{fig:unlinkability}
\vspace{-0.55cm}
\end{figure}


\commentt{Using these distributions, two measures of unlinkability are specified: i) Local measure $D_{\leftrightarrow{}}(s)$ evaluates the linkability of the system for each specific linkage score $s$ and is dependent upon the likelihood ratio between score distributions. $D_{\leftrightarrow{}}(s) \in [0,1]$ and is defined over the entire score domain. $D_{\leftrightarrow{}}(s) = 0$ denotes full unlinkability, while $D_{\leftrightarrow{}}(s) = 1$ denotes full linkability of two transformed templates at score $s$. All values of $D_{\leftrightarrow{}}(s)$ between 0 and 1 indicate an increasing degree of linkability. ii) Global measure $D_{\leftrightarrow{}}^{sys}$ provides an overall measure of the linkability of the system independent of the score domain and is a fairer benchmark for unlinkability comparison of two or more systems. $D_{\leftrightarrow{}}^{sys} \in [0,1]$, where $D_{\leftrightarrow{}}^{sys}=1$ indicates full linkability for all the scores of the mated samples distribution and  $D_{\leftrightarrow{}}^{sys}=1$ indicates full unlinkability for the whole score domain. All values of $D_{\leftrightarrow{}}^{sys}$ between 0 and 1 indicates an increasing degree of linkability.}

\commentt{According to the benchmark protocol defined in \cite{unlinkability_analysis}, six transformed databases were generated from WVU Multimodal face and iris test dataset by using different set of random bits (enrollment key) in the CTM for each template of a subject. The linkage score we have used is the Hamming distance between the $\textbf{s}_\textbf{e}$ and $\textbf{s}_\textbf{p}$. The mated samples distribution and the non-mated samples distribution were computed across these six databases. These score distributions are used to calculate local measure $D_{\leftrightarrow{}}(s)$, which is further used to compute the global measure $D_{\leftrightarrow{}}^{sys}$ (overall linkability of the system). Fig. \ref{fig:unlinkability} shows
unlinkability curves when transformed templates are generated for joint-FCA, and joint-BLA modalities using $m=8, N'=255$, and $n=768$. We have tested with two quantities of security bits $k=104$ and $k=128$  bits. With significant overlap, the overall linkability of the system is close to zero for both joint-FCA ($D_{\leftrightarrow{}}^{sys}=0.048$) and joint-BLA ($D_{\leftrightarrow{}}^{sys}=0.038$). Based on this discussion, the proposed system can be considered to be unlinkable.}

\vspace{-0.20cm}
\section{Conclusion}\label{sec:conc}

We have presented a feature-level fusion and binarization framework using deep hashing to design a multimodal template protection scheme that generates a single secure template from each user's multiple biometrics. We have employed a hybrid secure architecture combining the secure primitives of \emph{cancelable biometrics} and \emph{secure-sketch} and integrated it with a deep hashing framework, which makes it computationally prohibitive to forge a combination of multiple biometrics that passes the authentication. We have also proposed two  deep learning based fusion architectures, \emph{fully connected architecture} and \emph{bilinear architecture} that could be used to combine more than two modalities. Moreover, we have analyzed the matching performance and the security, and also performed also unlinkability analysis of the proposed secure multibiometric system. Experiments using the WVU multimodal dataset, which contain face and iris modalities, demonstrate that the matching performance does not deteriorate with the proposed protection scheme. In fact, both the matching performance and the template security are improved when using the proposed secure multimodal system. \commentAQ{However, we want to clarify that while the proposed solution is an interesting biometric security framework, in particular for structured data from modalities like face and iris, further validation is required to show how much it can work with other biometric modalities. Finally, the goal of this paper is to motivate researchers to investigate how to generate secure compact multimodal  templates.}

\vspace{-0.45cm}
\comment{\appendix[Image-retrieval Efficiency on ImageNet Dataset]
In order to test the effectiveness of the hashing layer in our proposed methods, we have also tested our deep hashing method for image retrieval on the ImageNet (ILSVRC 2015) \cite{ILSVRC15} dataset and compared the retrieval performance against some baseline hashing methods. The ImageNet dataset contains over 1.2 million images in the training set and about 50 thousand images in the validation set corresponding to 1000 categories. For comparison, we follow the same setting in \cite{cao_2017_hashnet}. We randomly select 100 categories and use all the corresponding training set images as our database and corresponding validation set images as our query points. We select 100 images per category from database as training points.}

\begin{table}[t]
\centering
\captionsetup{width=.95\linewidth}
\caption{\comment{Mean average precision (MAP $\%$) comparison with other hashing methods for 32, 48 and 64 bits.}}
\scalebox{0.75}{\begin{tabular}{cccc}
 \hline
\multicolumn{1}{c}{\multirow{1}{*}{Methods}} &\multicolumn{3}{c}{\multirow{1}{*}{ImageNet}} \\ [0.5ex] 
 \cline{2-4}
 &32 &48& 64  \\ \hline 
 LSH\cite{gionis_1999_similarity}&25.42&33.74&36.18 \\
 ITQ\cite{iterative_gong_2013}&46.96&53.23&57.05 \\
 CCA-ITQ\cite{iterative_gong_2013}&47.1&55.67&58.80 \\
 DHN\cite{zhu_2016_deep}&49.17&57.19&59.82 \\
 HashNet\cite{cao_2017_hashnet}&\textbf{63.48}&66.07&68.51 \\
 SSDH\cite{ssdh_2018_pami}&63.26&66.34&68.68 \\
 Our Hashing Method &63.18&\textbf{66.85}&\textbf{69.12} \\ \hline
\end{tabular}}
\label{table:MAP}
\vspace{-0.35cm}
\end{table}


\comment{ For evaluation, we use Mean Average Precision (MAP@1000), Precision curves with Hamming radius 2 ($P@r=2$), and Precision curves for different numbers of top returned samples ($P@K$). We compare our proposed hashing method with 6 state-of-the-art hashing methods including shallow hashing methods LSH \cite{gionis_1999_similarity}, ITQ \cite{iterative_gong_2013}, CCA-ITQ \cite{iterative_gong_2013}, and the deep hashing methods DHN \cite{zhu_2016_deep}, HashNet \cite{cao_2017_hashnet} and SSDH\cite{ssdh_2018_pami}. We report results using source code provided by the respective authors except for DHN for which we report result published in \cite{cao_2017_hashnet}.  For all the shallow hashing methods, we use VGG-19 fc7 features as input, and for deep hashing methods, we use raw images as input. For fair comparison we use VGG-19 for all the deep hashing methods.}

\comment{We can observe from the MAP comparison in Table \ref{table:MAP} that our hashing technique is better than shallow hashing methods for all hash code lengths. Also, our hashing method is competitive with the other state-of-the-art deep hashing methods when the size of the hash code is 32 bits, at the higher hash code lengths of 48 and 64 bits, our hashing technique is slightly better than other deep hashing methods by $0.35\%$.}

\begin{figure}[t]
\centering    
\subfigure[P@$r=2$]{\label{subfig:map_radius}\includegraphics[width=4.75cm]{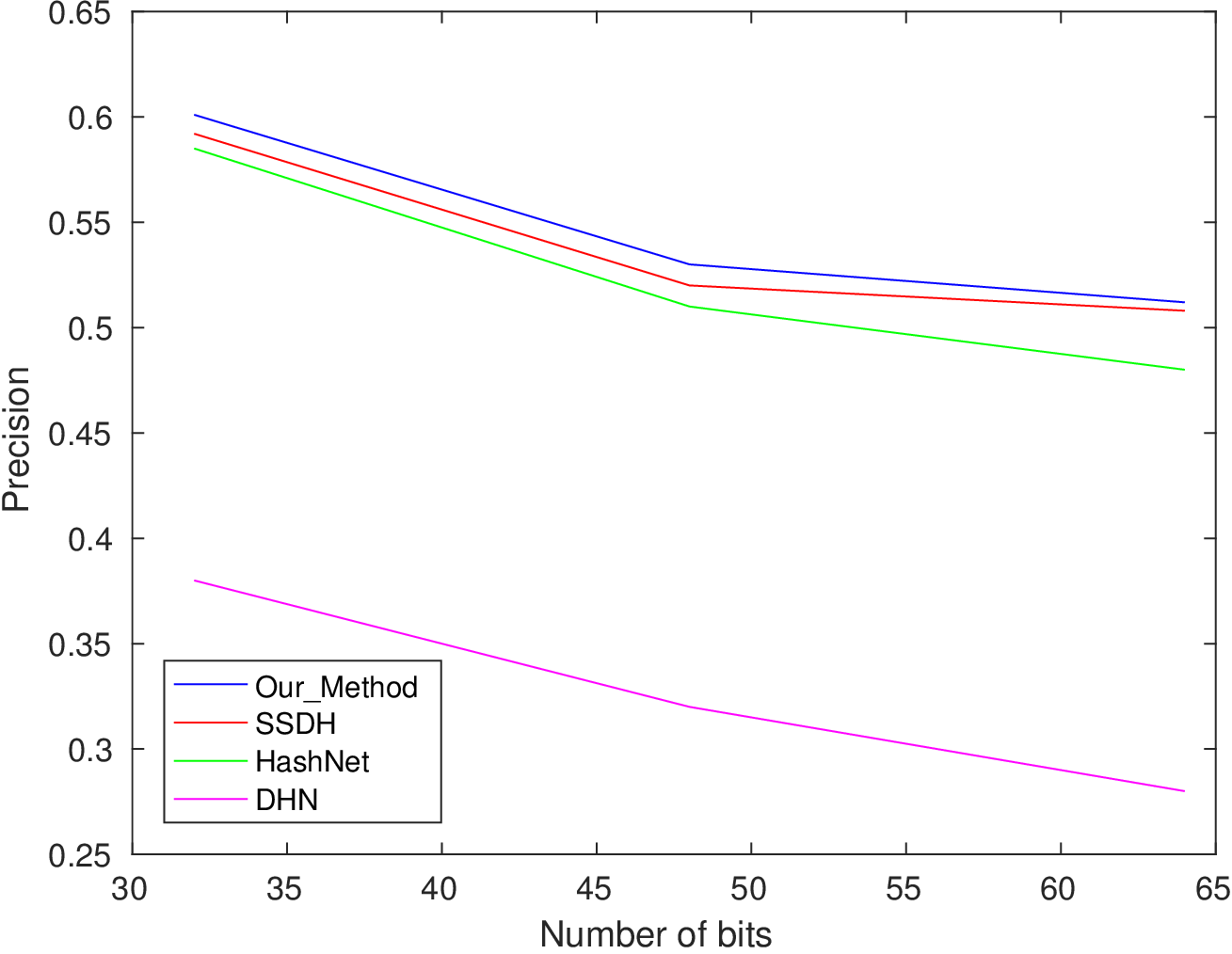}}
\subfigure[P@K for 64 bits]{\label{subfig:map_top_results}\includegraphics[width=4.75 cm]{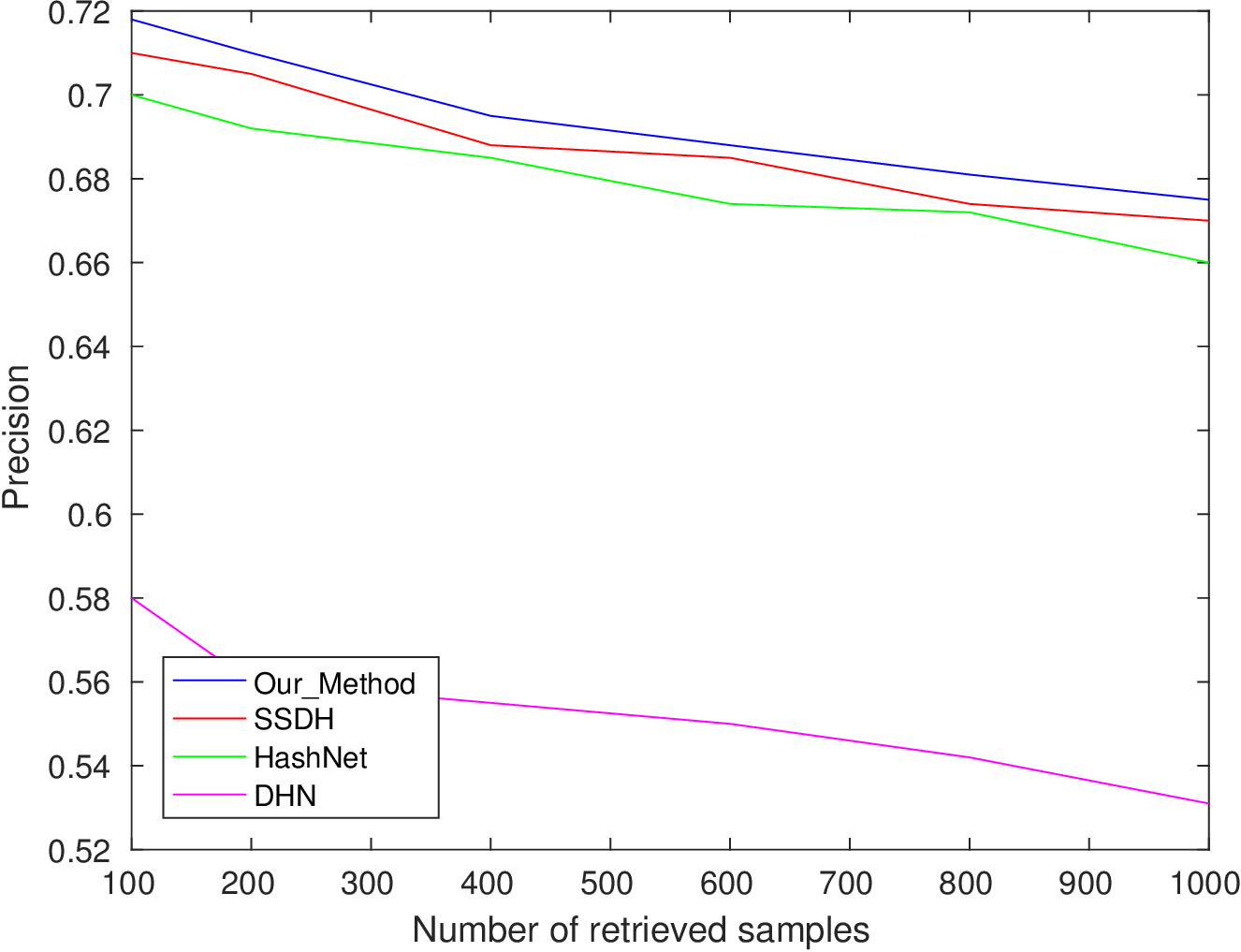}}
\caption{\comment{Experimental precision results for the ImageNet dataset for different deep hashing methods.}}
\label{fig:MAP}
\vspace{-0.45cm}
\end{figure}

\comment{Fig. \ref{subfig:map_radius} shows the Hamming precision curves for Hamming radius $r=2$ (P@$r=2$) for different hash code lengths only for the deep hashing methods. Fig. \ref{subfig:map_top_results} shows the precision for hash code length of 64 bits for different number of top retrieved results (P@K) only for the deep hashing methods. Our hashing technique consistently provides better precision than all the other hashing methods for the same number of retrieved results. Also, it is noted from Fig. \ref{subfig:map_radius} that precision at 32 bits is better than the precision at 48 and 64 bits. This is because  when using longer binary codes, the data distribution in Hamming space becomes progressively sparse and fewer samples fall within the set Hamming radius \cite{Yuan_2018_ECCV}.}

\vspace{-0.20cm}
\section*{Acknowledgment}
This research was funded by the Center for Identification Technology Research (CITeR), a National Science Foundation (NSF) Industry/University Cooperative Res. Center (I/UCRC).
\vspace{-0.45cm}

\ifCLASSOPTIONcaptionsoff
  \newpage
\fi



\balance
{\small
\bibliographystyle{IEEEtran}
\bibliography{deep_multimodal}
}

%



%






\vspace{-0.75cm}
\begin{IEEEbiography}[{\includegraphics[width=1in,height=1.50in,clip,keepaspectratio]{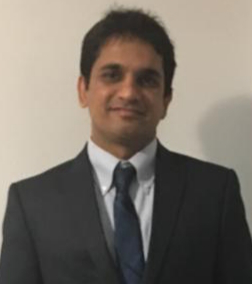}}]{Veeru Talreja} is a Ph.D. candidate at West Virginia  University (WVU),  Morgantown,  WV, USA.  He received the  M.S.E.E. degree from  West Virginia  University and the B.Engg. degree from Osmania University, Hyderabad, India. From 2010 to 2013 he worked as a Geospatial Software Developer with West Virginia  University research corporation. His  research  interests include applied machine learning, deep learning, coding theory, multimodal biometric recognition and security, and image retrieval.  
\end{IEEEbiography}

 \vspace{-0.65cm}
\begin{IEEEbiography}[{\includegraphics[width=1in,height=1.25in,clip,keepaspectratio]{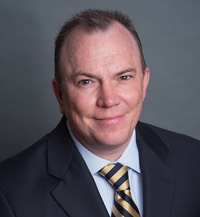}}]{Matthew  C.  Valenti} (M’92  -  SM’07  -  F’18)  received the M.S.E.E. degree from the Johns Hopkins University, Baltimore, MD, USA, and B.S.E.E. and Ph.D. degrees from Virginia Tech, Blacksburg, VA,USA.  He  has  been  a  Faculty  Member  with  West Virginia University since 1999, where he is currently a Professor and the Director of the Center for Identification Technology Research. His research interests are  in  wireless  communications,  cloud  computing, and biometric identification. He is the recipient of the 2019 MILCOM Award for Sustained Technical Achievement. He is active in the organization and oversight of several ComSoc sponsored IEEE conferences, including MILCOM, ICC, and Globecom. He was Chair of the  ComSoc Communication  Theory  Technical  committee  from  2015-2016, was  TPC  chair  for  MILCOM’17,  is  Chair  of  the  Globecom/ICC  Technical Content  (GITC)  Committee  (2018-2019), and  is  TPC  co-chair  for  ICC’21 (Montreal).  He  was  an  Electronics  Engineer  with  the  U.S.  Naval  Research Laboratory, Washington, DC, USA. Dr. Valenti is registered as a Professional Engineer in the state of West Virginia
\end{IEEEbiography}
\vspace{-0.65cm}
\begin{IEEEbiography}[{\includegraphics[width=1.25in,height=1.25in,clip,keepaspectratio]{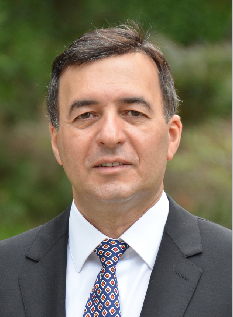}}]{Nasser M. Nasrabadi} (S’80 – M’84 – SM’92 – F’01) received the B.Sc. (Eng.) and Ph.D. degrees in electrical engineering from the Imperial College of Science and Technology, University of London, London, U.K., in 1980 and 1984, respectively. In 1984, he was with IBM, U.K., as a Senior Programmer. From 1985 to 1986, he was with the Philips Research Laboratory, New York, NY, USA, as a member of the Technical Staff. From 1986 to 1991, he was an Assistant Professor with the Department of Electrical Engineering, Worcester Polytechnic Institute, Worcester, MA, USA. From 1991 to 1996, he was an Associate Professor with the Department of Electrical and Computer Engineering, State University of New York at Buffalo, Buffalo, NY, USA. From 1996 to 2015, he was a Senior Research Scientist with the U.S. Army Research Laboratory. Since 2015, he has been a Professor with the Lane Department of Computer Science and Electrical Engineering. His current research interests are in image processing, computer vision, biometrics, statistical machine learning theory, sparsity, robotics, and neural networks applications to image processing. He is a fellow of ARL and SPIE and has served as an Associate Editor for the IEEE Transactions on Image Processing, the IEEE Transactions on Circuits, Systems and Video Technology, and the IEEE Transactions on Neural Networks.
\end{IEEEbiography}

\end{document}